\documentclass{article}

\usepackage[preprint]{neurips_2019}

\usepackage[utf8]{inputenc} % allow utf-8 input
\usepackage[T1]{fontenc}    % use 8-bit T1 fonts
\usepackage{hyperref}       % hyperlinks
\usepackage{url}            % simple URL typesetting
\usepackage{booktabs}       % professional-quality tables
\usepackage{amsfonts}       % blackboard math symbols
\usepackage{nicefrac}       % compact symbols for 1/2, etc.
\usepackage{microtype}      % microtypography

\usepackage{float}
\usepackage{subfig}
\usepackage{graphicx}
\usepackage{multirow}
\usepackage{framed}
\usepackage{amsmath}
\usepackage{svg}

\title{An Efficient Graph Convolutional Network Technique\\ for the Travelling Salesman Problem}

\author{
  Chaitanya K. Joshi$^{1}$, 
  Thomas Laurent$^{2}$, 
  and Xavier Bresson$^{1}$ \\
  $^{1}$School of Computer Science and Engineering, 
  Nanyang Technological University, Singapore \\
  $^{2}$Department of Mathematics, 
  Loyola Marymount University \\
  \texttt{\{chaitanya.joshi, xbresson\}@ntu.edu.sg, tlaurent@lmu.edu} \\
}

\begin{document}

\maketitle

\begin{abstract}

This paper introduces a new learning-based approach for approximately solving the Travelling Salesman Problem on 2D Euclidean graphs.
We use deep Graph Convolutional Networks to build efficient TSP graph representations and output tours in a non-autoregressive manner via highly parallelized beam search.
Our approach\footnote{Code and data available at \url{https://github.com/chaitjo/graph-convnet-tsp}.} 
outperforms all recently proposed autoregressive deep learning techniques in terms of solution quality, inference speed and sample efficiency for problem instances of fixed graph sizes. 
In particular, we reduce the average optimality gap from $0.52\%$ to $0.01\%$ for 50 nodes, and from $2.26\%$ to $1.39\%$ for 100 nodes. 
Finally, despite improving upon other learning-based approaches for TSP, our approach falls short of standard Operations Research solvers.
\end{abstract}

\section{Introduction}
\label{introduction}

NP-hard combinatorial optimization problems are the family of integer constrained optimization problems which are intractable to solve  optimally at large scales. 
%In practice it is possible to approximate solutions for instances up to a million decision variables and constraints \citep{applegate2006traveling}. 
Robust approximation algorithms to popular NP-hard problems have various practical applications and are the backbone of modern industries such as transportation, supply chain, energy, finance, and scheduling. 

One of the most famous NP-hard problems, the Travelling Salesman Problem (TSP), asks the following question: ``Given a list of cities and the distances between each pair of cities, what is the shortest possible route that visits each city and returns to the origin city?" 
Formally, given a graph, one needs to search the space of permutations to find an optimal sequence of nodes, called a tour, with minimal total edge weights (tour length). 
In general, NP-hard problems can be formulated as sequential decision making tasks on graphs due to their highly structured nature. 
Thus, machine learning can be used to train policies for approximately solving these problems instead of handcrafting solutions, which may be expensive or require significant specialized knowledge \citep{bengio2018machine}. 
In particular, recent advances in graph neural network techniques \citep{bruna2013spectral,defferrard2016convolutional,sukhbaatar2016gcn,kipf2016semi,hamilton2017inductive} are a good fit for the task because they naturally operate on the graph structure of these problems.

Recently proposed deep learning approaches for the 2D Euclidean TSP combine graph neural networks with autoregressive decoding to output TSP tours one node at a time using the \textit{sequence-to-sequence} framework \citep{vinyals2015pointer,bello2016neural} or an attention mechanism \citep{deudon2018learning,kool2018attention}. 
Policies are trained using reinforcement learning where the partial tour length is used to formulate a reward function at each step.

\begin{figure}[t!]
    \includegraphics[width=0.95\textwidth]{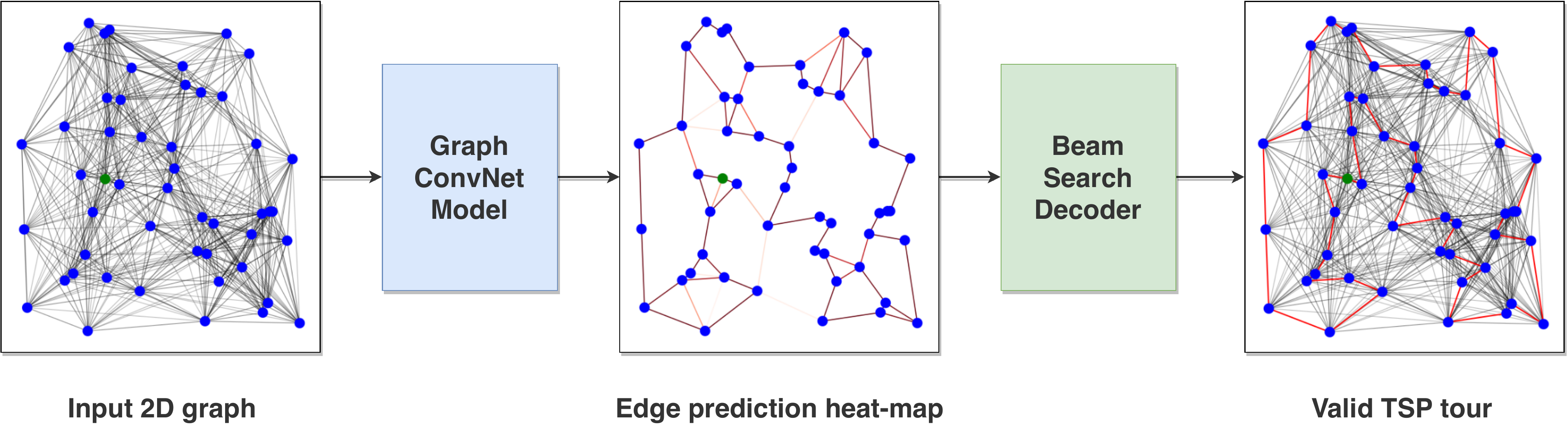}
    \caption{Overview of our approach. Taking a 2D graph as input, the graph ConvNet model outputs an edge adjacency matrix denoting the probabilities of edges occurring on the TSP tour. This is converted to a valid tour using beam search. All components are highly parallelized and solutions are produced in a one-shot, non-autoregressive manner.}
    \label{fig:blocks}
\end{figure}

In this paper, we introduce a non-autoregressive deep learning approach for approximately solving TSP using the \textit{Graph Convolutional Network} (graph ConvNet) introduced in \citep{bresson2017residual} and the beam search technique \citep{medress1977speech}. 
Figure \ref{fig:blocks} presents an overview of our approach. Our model takes a graph as an input and extracts compositional features from its nodes and edges by stacking several graph convolutional layers. 
The output of the neural network is an edge adjacency matrix denoting the probabilities of edges occurring on the TSP tour. 
The edge predictions, forming a \textit{heat-map}, are converted to a valid tour using a post-hoc beam search technique. 
The model parameters are trained in a supervised manner using pairs of problem instances and optimal solutions using the Concorde TSP solver \citep{applegate2006traveling}. 

We demonstrate the efficiency and speed of our approach over other deep learning techniques through empirical comparisons on TSP instances of fixed graph sizes with 20, 50 and 100 nodes:
\begin{itemize}
    \item \textbf{Solution quality:} We efficiently train deep graph ConvNets with better representation capacity compared to previous approaches, leading to significant gains in solution quality (in terms of closeness to optimality).
    \item \textbf{Inference speed:} Our graph ConvNet and beam search implementations are highly parallelized for GPU computation, leading to fast inference time and better scalability to large graphs. 
    In contrast, autoregressive approaches scale poorly to large graphs due to the sequential nature of the decoding process, which cannot be parallelized. 
    \item \textbf{Sample efficiency:} Our supervised training setup using pairs of problem instances and optimal solutions is more sample efficient compared to reinforcement learning. We are able to learn better approximate solvers using lesser training data.
\end{itemize}

\section{Related Work}
\label{related}

The Traveling Salesman Problem (TSP), first formulated in 1930, is one of the most intensively studied combinatorial optimization problems in the Operations Research (OR) community.
Finding the optimal TSP solution is NP-hard, even in the 2D Euclidean case where the nodes are 2D points and edge weights are Euclidean distances between pairs of points \citep{papadimitriou1977euclidean}. 
In practice, TSP solvers rely on carefully handcrafted heuristics to guide their search procedures for finding approximate solutions efficiently for graphs with thousands of nodes.
Today, state-of-the-art TSP solvers such as Concorde \citep{applegate2006traveling} make use of cutting plane algorithms \citep{dantzig1954solution,padberg1991branch,applegate2003implementing} to iteratively solve linear programming relaxations of the TSP in addition to a branch-and-bound approach that reduces the solution search space.

Thus, designing good heuristics for combinatorial optimization problems often requires significant specialized knowledge and years of research work.
Due to the highly structured nature of these problems, 
neural networks have been used to learn approximate policies instead, especially for problems that are non-trivial to design heuristics for \citep{smith1999neural,bengio2018machine}.
Historical work has focused on learning-based approaches for TSP using Hopfield networks \citep{hopfield1985neural} 
and deformable template models \citep{fort1988solving,angeniol1988self}.
However, benchmark performance for these approaches has not matched algorithmic methods in terms of speed and solution quality \citep{la2012comparison}.

Recent advances in \textit{sequence-to-sequence} learning \citep{sutskever2014sequence}, attention mechanisms \citep{bahdanau2014neural} and geometric deep learning \citep{bronstein2017geometric} have reinvigorated this line of work.  
\cite{vinyals2015pointer} introduced the sequence-to-sequence \textit{Pointer Network} (PtrNet) model that uses attention to output a permutation of an input sequence.
The model is trained to autoregressively output TSP tours in a supervised manner via pairs of problem instances and solutions generated by Concorde.
Upon test time, they use a beam search procedure to build valid tours in a fashion similar to neural machine translation \citep{wu2016google}.
\citet{bello2016neural} trained the PtrNet without supervised solutions by using an Actor-Critic reinforcement learning algorithm. 
They consider each instance as a training sample and use the cost (tour length) of a sampled solution for an unbiased Monte-Carlo estimate of the policy gradient.

\cite{dai2017learning} encoded problem instances using graph neural networks, which are invariant to node order and better reflect the combinatorial structure of TSP compared to sequence-to-sequence models.
They train a \texttt{structure2vec} graph embedding model \citep{dai2016discriminative} to output the order in which nodes are inserted into a partial tour using the DQN training method \citep{mnih2013playing} and a helper function to insert at the best possible location. 

Concurrent work by \cite{deudon2018learning} and \cite{kool2018attention} replaced the \texttt{structure2vec} model with the recently proposed \textit{Graph Attention Network} \citep{velivckovic2017graph} and used an attention-based decoder trained with reinforcement learning to autoregressively build TSP solutions. 
\cite{deudon2018learning} showed that a hybrid approach of using 2OPT local search \citep{croes1958method} on top of tours produced by the model improves performance.
\cite{kool2018attention} used a more powerful decoder and trained the model using REINFORCE \citep{williams1992simple} with a greedy rollout baseline to achieve state-of-the-art results among learning-based approaches for TSP.

In contrast to autoregressive approaches, \cite{nowak2017note} trained a graph neural network \citep{scarselli2009graph} in a supervised manner to directly output a tour as an adjacency matrix, which is converted into a feasible solution using beam search. 
Due to its one-shot nature, the model cannot condition its output on the partial tour and performs poorly for very small problem instances.
Our non-autoregressive approach builds on top of this work.
% Our non-autoregressive approach builds on top of this work by using graph ConvNets \citep{bresson2017residual} and a weighted loss function to scale up to large problem instances.

Similar learning-based techniques have been proposed for generalizations of TSP such as the Vehicle Routing Problem \citep{nazari2018reinforcement,kool2018attention} and the multiple TSP \citep{kaempfer2018learning}, as well as other combinatorial problems such as the Minimum Vertex Cover, Maximum Cut and Maximal Independent Set \citep{dai2017learning,li2018combinatorial,venkatakrishnan2018graph2seq,mittal2019learning}.

% Table \ref{table:model-types} summarizes recent learning-based approaches to TSP, starting from \cite{vinyals2015pointer}. 

% \begin{table}[h!]
% \centering
% \caption[Recent learning-based approaches to TSP.]{Recent learning-based approaches to TSP characterized by type of model, training setting and solution search. In the \emph{Training Setting} column, \textbf{SL} denotes Supervised Learning and \textbf{RL} denotes Reinforcement Learning.}
% \label{table:model-types}
% \resizebox{\textwidth}{!}{%
% \begin{tabular}{lcccc}
% \toprule
% Method & Neural Network & Model Type & Training Setting & Solution Search Type \\
% \midrule
% \cite{vinyals2015pointer} & Pointer Network & Autoregressive & SL & Beam search \\
% \cite{bello2016neural} & Pointer Network & Autoregressive & RL & Sample from policy \\
% \cite{dai2017learning} & Structure2vec & Autoregressive & RL & Greedy algorithm \\
% \cite{nowak2017note} & Graph Neural Network & Non-autoregressive & SL & Beam search \\
% \cite{deudon2018learning} & Graph Attention Network & Autoregressive & RL & Sample from policy \\
% \cite{kool2018attention} & Graph Attention Network & Autoregressive & RL & Sample from policy \\
% Ours & Graph ConvNet & Non-autoregressive & RL & Beam search \\ 
% \bottomrule
% \end{tabular}%
% }
% \end{table}

\section{Dataset}
\label{data}

We focus on the 2D Euclidean TSP, although the presented technique can also be applied to sparse graphs. 
Given an input graph, represented as a sequence of $n$ cities (nodes) in the two dimensional unit square $S = \{x_i\}_{i=1}^n$ where each $x_i \in {[0,1]}^2$, we are concerned with finding a permutation of the points $\hat \pi$, termed a tour, that visits each node once and has the minimum total length. We define the length of a tour defined by a permutation $\hat \pi$ as
\begin{equation}
L(\hat{\pi} | s) = \|\mathbf{x}_{\hat \pi(n)} - \mathbf{x}_{\hat \pi(1)}\|_2
    + \sum_{i=1}^{n-1} \|\mathbf{x}_{\hat \pi(i)} - \mathbf{x}_{\hat \pi(i+1)}\|_2
\end{equation}
where $\|\cdot\|_2$ denotes the $\ell_2$ norm. 

Introduced by \cite{vinyals2015pointer}, the current paradigm for learning-based approaches to TSP is based on training and evaluating model performance on problem instances of fixed sizes. 
Hence, we create separate training, validation and test datasets for graphs of sizes 20, 50 and 100 nodes. 
The training sets consists of one million pairs of problem instances and solutions, and the validation and test sets consist of 10,000 pairs each. For each TSP instance, the $n$ node locations are sampled uniformly at random in the unit square. The optimal tour $\pi$ is found using Concorde \citep{applegate2006traveling} \footnote{Code available at \url{http://www.math.uwaterloo.ca/tsp/concorde.html}.}.
See Appendix \ref{dataset-stats} for dataset summary statistics.

\section{Model}
\label{model}

Given a graph as an input, we train a graph ConvNet model to directly output an adjacency matrix corresponding to a TSP tour. The network computes $h$-dimensional representations for each node and edge in the graph. 
The edge representations are linked to the ground-truth TSP tour through a softmax output layer so that the model parameters can be trained \textit{end-to-end} by minimizing the cross-entropy loss via gradient descent. During test time, the adjacency matrix obtained from the model is converted to a valid tour via beam search.

\subsection{Graph ConvNet}
\label{gcn}

\paragraph{Input layer} As input node feature, we are given the two dimensional coordinates $x_i \in {[0,1]}^2$, which are embedded to $h$-dimensional features:
\begin{equation}
\label{eqn:node-input}
\alpha_i = A_1  x_i + b_1
\end{equation}
where $A_1 \in \mathbb{R}^{h \times 2} $. The edge Euclidean distance $d_{ij}$ is embedded as a $\frac{h}{2}$-dimensional feature vector. 
We also define an indicator function of a TSP edge $\delta^{\textrm{\tiny{k-NN}}}_{ij}$ with the value one if nodes $i$ and $j$ are $k$-nearest neighbors, value two for self-connections, and value zero otherwise. The edge input feature $\beta_{ij}$ is:
\begin{equation}
\label{eqn:edge-input}
\beta_{ij} = A_2  d_{ij} + b_2 \; \|\; A_3  \delta^{\textrm{\tiny{k-NN}}}_{ij} 
\end{equation}
where $A_2 \in \mathbb{R}^{\frac{h}{2} \times 1} $, 
$A_3 \in \mathbb{R}^{\frac{h}{2} \times 3} $, and 
$\cdot \| \cdot$ is the concatenation operator. The input $k$-nearest neighbor graph speeds up the learning process as a node in the TSP solution is usually connected to nodes in its close proximity.

\paragraph{Graph Convolution layer} Let  $x_i^{\ell}$ and $e_{ij}^{\ell}$ denote respectively the node feature vector and edge feature vector at layer $\ell$ associated with node $i$ and edge $ij$. We define the node feature and edge feature at the next layer as:
\begin{eqnarray}
x_i^{\ell+1} &=& x_i^{\ell} + \text{ReLU} \Big( \text{BN} \Big( W_1^\ell x_i^{\ell} + \sum_{j\sim i} \eta_{ij}^{\ell} \odot W_2^\ell x_j^{\ell} \Big)\Big) \ \textrm{with} \ \eta_{ij}^{\ell} = \frac{\sigma(e_{ij}^{\ell})}{\sum_{j'\sim i} \sigma(e_{ij'}^{\ell}) + \varepsilon }, \label{eqn:gcn-node}\\
e_{ij}^{\ell+1} &=& e^\ell_{ij} + \text{ReLU}\Big( \text{BN} \Big( W_3^\ell e_{ij}^{\ell} + W_4^\ell x^{\ell}_i + W_5^\ell x^{\ell}_j  \Big)\Big),\label{eqn:gcn-edge}
\end{eqnarray}
where $W \in \mathbb{R}^{h \times h} $, $\sigma$ is the sigmoid function, $\varepsilon$ is a small value, $\text{ReLU}$ is the rectified linear unit, and $\text{BN}$ stands for batch normalization. 
At the input layer, we have $x_i^{\ell=0}=\alpha_i$ and $e_{ij}^{\ell=0}=\beta_{ij}$. The proposed graph ConvNet leverages \cite{bresson2017residual} with the additional edge feature representation and a dense attention map $\eta_{ij}^{\ell}$, which makes the diffusion process anisotropic on graphs.

See Appendix \ref{model-details} for a detailed description of the graph convolution layer.

\paragraph{MLP classifier}
The edge embedding $e_{ij}^{\ell}$ of the last layer is used to compute the probability of that edge being connected in the TSP tour of the graph. This probability can be seen as computing a probabilistic heat-map over the adjacency matrix of tour connections. Each $p^{\textrm{\tiny{TSP}}}_{ij} \in {[0,1]}^2$ is given by a \textit{Multi-layer Perceptron} (MLP):
\begin{equation}
\label{eqn:mlp}
p^{\textrm{\tiny{TSP}}}_{ij} = \textrm{MLP}(e_{ij}^{L})
\end{equation}
In practice, we may have an arbitrary number of layers denoted by $l_{mlp}$.

\paragraph{Loss function}
Given the ground-truth TSP tour permutation $\pi$, we convert the tour into an adjacency matrix where each element $\hat{p}_{ij}^{\textrm{\tiny{TSP}}}$ denotes the presence or absence of an edge between nodes $i$ and $j$ in the TSP tour. We minimize the weighted binary cross-entropy loss averaged over mini-batches.
As the problem size increases, the classification task becomes highly unbalanced towards the negative class, which requires appropriate class weights to balance this effect.
\footnote{We compute balanced class weights for each instance of TSP$n$ as $w_0=\frac{n^2}{(n^2 - 2n)\times c}$ and $w_1=\frac{n^2}{(2n)\times c}$, where $c = 2$ denotes the number of classes.}

\subsection{Beam Search Decoding}
\label{decoding}

The output of our model is a probabilistic heat-map over the adjacency matrix of tour connections. Each $p^{\textrm{\tiny{TSP}}}_{ij} \in {[0,1]}^2$ denotes the strength of the edge prediction between nodes $i$ and $j$. 
Based on the chain rule of probability, the probability of a partial TSP tour $\pi'$ can be formulated as:
\begin{equation}
p(\pi') = \prod_{j' \sim i' \in \pi'} p^{\textrm{\tiny{TSP}}}_{i'j'}
\end{equation}
where each node $j'$ follows node $i'$ in the partial tour $\pi'$.
However, directly converting the probabilistic heat-map to an adjacency matrix representation of the predicted TSP tour $\hat \pi$ via an $\arg\!\max$ function will generally yield invalid tours with extra or not enough edges in $\hat \pi$.  
Thus, we employ three possible search strategies at evaluation time to convert the probabilistic edge heat-map to a valid permutation of nodes $\hat \pi$.

\paragraph{Greedy search} In general, greedy algorithms choose the local optimal solution to provide a fast approximation of the global optimal solution. Starting from the first node, we greedily select the next node from its neighbors based on the highest probability of the presence of an edge. The search terminates when all nodes have been visited. We mask out nodes that have previously been visited in order to construct valid solutions.

% It is interesting to draw a parallel between our probabilistic greedy search and a popular distance-based greedy strategy for TSP \citep{gutin2002traveling}. The distance-based heuristic is described as: ``At each step of the journey, visit the nearest unvisited city." Our greedy search can similarly be described as: ``At each step of the journey, take the most probable path to an un-visited city."

\paragraph{Beam search} 
A beam search is a limited-width breadth-first search \citep{medress1977speech}. 
Beam search is a popular approach for obtaining a set of high-probability sequences from generative models for natural language processing tasks \citep{wu2016google}. 
% Starting from an empty sequence (e.g. $t = 0$), a beam search expands at every step $t = 0, 1, 2, ...$ at most $b$ partial sequences (those with highest probability) to compute the probabilities of sequences with length $t + 1$. It terminates with a beam of $b$ complete sequences. (Note that $b$ is referred to as the beam width.)
Starting from the first node, we explore the heat-map by expanding the $b$ most probable edge connections among the node's neighbors. 
We iteratively expand the top-$b$ partial tours at each stage till we have visited all nodes in the graph. 
We follow the same masking strategy as greedy search to construct valid tours. 
The final prediction is the tour with the highest probability among the $b$ complete tours at the end of beam search. 
(Note that $b$ is referred to as the beam width.)
% Other supervised learning-based techniques for TSP \citep{vinyals2015pointer,nowak2017note} follow the same search strategy to generate valid solutions. 

\paragraph{Beam search and Shortest tour heuristic} 
Instead of selecting the tour with the highest probability at the end of beam search, we select the shortest tour among the set of $b$ complete tours as the final solution. 
This heuristic-based beam search is directly comparable to reinforcement learning techniques for TSP which sample a set of solutions from the learned policy and select the shortest tour among the set as the final solution \citep{bello2016neural,kool2018attention}.

\subsection{Hyperparameter Configurations}

We use an identical set of model hyperparameters across all three problem sizes.
Each model consists of $l_{conv} = 30$ graph convolutional layer and $l_{mlp} = 3$ layers in the MLP with hidden dimension $h = 300$ for each layer. 
We use a fixed beam width $b = 1,280$ in order to directly compare our results to the current state-of-the-art \citep{kool2018attention} which samples 1,280 solutions from a learned policy.
We consider $k = 20$ nearest neighbors for each node in the adjacency matrix $W^{\textrm{\tiny{adj}}}$.
(We found the $20$-nearest neighbor graph to be an approximate upper bound on the TSP solution space for the training sets of each problem size.)

\section{Experiments}
\label{experiments}

\subsection{Training Procedure}

We follow a standard training procedure to train models for each problem size. Given a graph as an input, we train the graph ConvNet model to directly output an adjacency matrix corresponding to a TSP tour by minimizing the cross-entropy loss via gradient descent. 

\paragraph{Training loop}
For each training epoch, we randomly select a subset of 10,000 problem instances out of one million from the training set. 
The subset is divided into 500 mini-batches of 20 instances each. We use the Adam optimizer \citep{kingma2014adam} with an initial learning rate of $0.001$ to minimize the cross-entropy loss over each mini-batch.

\paragraph{Learning rate decay}
We evaluate our model on a held-out validation set of 10,000 instances at regular intervals of five training epochs.
If the validation loss has not decreased by at least 1\% of the previous validation loss, we divide the optimizer's learning rate by a decay factor of 1.01.
Using smaller learning rates as training progresses allows our models to learn faster and converge to better local minima.
% Larger problem instances require more training epochs and lower learning rates to reach convergence.

\subsection{Evaluation Procedure}
 
During evaluation on the validation and test sets, the adjacency matrix obtained from the model is converted to a valid tour via search strategies described in Section \ref{decoding}. 
As we do not need to do backpropagation during evaluation, we use arbitrarily large batch sizes that fit the entire GPU memory. 
Following \citep{kool2018attention}, we report the following metrics to evaluate performance of our models compared to optimal solutions (obtained using Concorde):

\paragraph{Predicted tour length} The average predicted TSP tour length $l^{\textrm{\tiny{TSP}}}$ over 10,000 test instances, computed as $\frac{1}{m} \sum_{i=1}^m l^{\textrm{\tiny{TSP}}}_m$.

\paragraph{Optimality gap} The average percentage ratio of the predicted tour length $l^{\textrm{\tiny{TSP}}}$ relative to the optimal solution $\hat{l}^{\textrm{\tiny{TSP}}}$ over 10,000 test instances, computed as 
$\frac{1}{m} \sum_{i=1}^m \left(  l_m^{\textrm{\tiny{TSP}}}/\hat{l}_m^{\textrm{\tiny{TSP}}} - 1 \right) $. 

\paragraph{Evaluation time} The total wall clock time taken to solve 10,000 test instances, either on a single GPU (Nvidia 1080Ti) or 32 instances in parallel on a 32 virtual CPU system (2 $\times$ Xeon E5-2630).

It is important to note that all deep learning approaches use search or sampling. Hence, it is possible to trade off run time for solution quality by  searching longer or sampling more solutions.
Run times can also vary due to implementations (Python vs C++) or hardware (GPU vs CPU).
\cite{kool2018attention} take a practical view and report the time it takes to solve the test set of 10,000 instances. 
% This is conservative: deep learning models are parallelizable while their non-learning baselines are single thread CPU implementations which cannot parallelize when running individually. 

\section{Results}
\label{results}

Table \ref{table:comparison} presents the performance of our technique compared to non-learned baselines and state-of-the-art deep learning techniques for various TSP instance sizes.
The table is divided into three sections: exact solvers, greedy methods (G), and sampling/search-based methods (S). Methods are further categorized according to the training technique: supervised learning (SL), reinforcement learning (RL), and non-learned heuristics (H).
All results except ours (in bold) are taken from Table 1 in \cite{kool2018attention}. 
More details about solvers (Concorde, LKH3 and Guorobi) and non-learned baselines (Nearest Insertion, Random Insertion, Farthest Insertion, Nearest Neighbor) can be found in Appendix B of \cite{kool2018attention}.
Following \cite{kool2018attention}, evaluation of TSP100 models on the test set was done with two GPUs and other timings were reported for single GPU models. 

\begin{table}[tb!]
\centering
\caption{Performance of our technique compared to non-learned baselines and state-of-the-art methods for various TSP instance sizes. 
Deep learning approaches are named according to the type of neural network used. 
The optimality gap is computed w.r.t Concorde. 
In the \emph{Type} column, \textbf{H}: Heuristic, \textbf{SL}: Supervised Learning, \textbf{RL}: Reinforcement Learning, \textbf{S}: Sampling, \textbf{G}: Greedy,  \textbf{BS}: Beam search, \textbf{BS*}: Beam search and shortest tour heuristic, and \textbf{2OPT}: 2OPT local search.
}
\label{table:comparison}
\resizebox{\textwidth}{!}{%
\begin{tabular}{lc|ccc|ccc|ccc}
\toprule
\multirow{2}{*}{Method} & \multirow{2}{*}{Type} & \multicolumn{3}{c|}{TSP20} & \multicolumn{3}{c|}{TSP50} & \multicolumn{3}{c}{TSP100} \\
 & & Tour Len. & Opt. Gap. & Time & Tour Len. & Opt. Gap. & Time & Tour Len. & Opt. Gap. & Time \\
\midrule
Concorde & Solver &  $3.84$ & $0.00 \%$ & (1m) & $5.70$ & $0.00 \%$ & (2m) & $7.76$ & $0.00 \%$ & (3m) \\
LKH3 & Solver &  $3.84$ & $0.00 \%$ & (18s) & $5.70$ & $0.00 \%$ & (5m) & $7.76$ & $0.00 \%$ & (21m) \\
Gurobi & Solver &  $3.84$ & $0.00 \%$ & (7s) & $5.70$ & $0.00 \%$ & (2m) & $7.76$ & $0.00 \%$ & (17m) \\
\midrule
Nearest Insertion & H, G &  $4.33$ & $12.91 \%$ & (1s) & $6.78$ & $19.03 \%$ & (2s) & $9.46$ & $21.82 \%$ & (6s) \\
Random Insertion & H, G &  $4.00$ & $4.36 \%$ & (0s) & $6.13$ & $7.65 \%$ & (1s) & $8.52$ & $9.69 \%$ & (3s) \\
Farthest Insertion & H, G &  $3.93$ & $2.36 \%$ & (1s) & $6.01$ & $5.53 \%$ & (2s) & $8.35$ & $7.59 \%$ & (7s) \\
Nearest Neighbor & H, G &  $4.50$ & $17.23 \%$ & (0s) & $7.00$ & $22.94 \%$ & (0s) & $9.68$ & $24.73 \%$ & (0s) \\
PtrNet \citep{vinyals2015pointer} & SL, G &  $3.88$ & $1.15 \%$ &  & $7.66$ & $34.48 \%$ &  & \multicolumn{3}{c}{-} \\
PtrNet \citep{bello2016neural} & RL, G  &  $3.89$ & $1.42 \%$ &  & $5.95$ & $4.46 \%$ &  & $8.30$ & $6.90 \%$ &  \\
S2V \citep{dai2017learning} & RL, G  &  $3.89$ & $1.42 \%$ &  & $5.99$ & $5.16 \%$ &  & $8.31$ & $7.03 \%$ &  \\
GAT \citep{deudon2018learning} & RL, G  &  $3.86$ & $0.66 \%$ & (2m) & $5.92$ & $3.98 \%$ & (5m) & $8.42$ & $8.41 \%$ & (8m) \\
GAT \citep{deudon2018learning} & RL, G, 2OPT  &  $3.85$ & $0.42 \%$ & (4m) & $5.85$ & $2.77 \%$ & (26m) & $8.17$ & $5.21 \%$ & (3h) \\
GAT \citep{kool2018attention} & RL, G  &  $3.85$ & $0.34 \%$ & (0s) & $5.80$ & $1.76 \%$ & (2s) & $8.12$ & $4.53 \%$ & (6s) \\
\textbf{GCN (Ours)} & \textbf{SL, G}  &  $\mathbf{3.86}$ & $\mathbf{0.60 \%}$ & \textbf{(6s)} & $\mathbf{5.87}$ & $\mathbf{3.10 \%}$ & \textbf{(55s)} & $\mathbf{8.41}$ & $\mathbf{8.38 \%}$ & \textbf{(6m)} \\
\midrule
OR Tools & H, S &  $3.85$ & $0.37 \%$ &  & $5.80$ & $1.83 \%$ &  & $7.99$ & $2.90 \%$ &  \\
Chr.f. + 2OPT & H, 2OPT  &  $3.85$ & $0.37 \%$ &  & $5.79$ & $1.65 \%$ &  & \multicolumn{3}{c}{-} \\
GNN \citep{nowak2017note} & SL, BS &  $3.93$ & $2.46 \%$ &  & \multicolumn{3}{c|}{-} & \multicolumn{3}{c}{-} \\
PtrNet \citep{bello2016neural} & RL, S  &  \multicolumn{3}{c|}{-} & $5.75$ & $0.95 \%$ &  & $8.00$ & $3.03 \%$ &  \\
GAT \citep{deudon2018learning} & RL, S &  $3.84$ & $0.11 \%$ & (5m) & $5.77$ & $1.28 \%$ & (17m) & $8.75$ & $12.70 \%$ & (56m) \\
GAT \citep{deudon2018learning} & RL, S, 2OPT  &  $3.84$ & $0.09 \%$ & (6m) & $5.75$ & $1.00 \%$ & (32m) & $8.12$ & $4.64 \%$ & (5h) \\
GAT \citep{kool2018attention} & RL, S  &  $3.84$ & $0.08 \%$ & (5m) & $5.73$ & $0.52 \%$ & (24m) & $7.94$ & $2.26 \%$ & (1h) \\
\textbf{GCN (Ours)} & \textbf{SL, BS}  &  $\mathbf{3.84}$ & $\mathbf{0.10 \%}$ & \textbf{(20s)} & $\mathbf{5.71}$ & $\mathbf{0.26 \%}$ & \textbf{(2m)} & $\mathbf{7.92}$ & $\mathbf{2.11 \%}$ & \textbf{(10m)} \\
\textbf{GCN (Ours)} & \textbf{SL, BS*}  &  $\mathbf{3.84}$ & $\mathbf{0.01 \%}$ & \textbf{(12m)} & $\mathbf{5.70}$ & $\mathbf{0.01 \%}$ & \textbf{(18m)} & $\mathbf{7.87}$ & $\mathbf{1.39 \%}$ & \textbf{(40m)} \\
\bottomrule
\end{tabular}%
}
\end{table}

\paragraph{Greedy setting}
In the greedy setting, learning-based approaches clearly outperform all non-learned heuristics.
% Deep learning approaches are able to solve TSP20 within 1.5\% of optimality, which suggests that the instance size is too simple for comparing models.
% For the larger instance sizes TSP50 and TSP100, graph neural network approaches \citep{deudon2018learning,kool2018attention} generally outperform sequence-to-sequence models \citep{vinyals2015pointer,bello2016neural}. 
Our graph ConvNet model is not able to match the performance or evaluation time of the GAT model of \cite{kool2018attention}. Indeed, autoregressive models are fast in this setting as they are specifically designed for it: they output the TSP tour permutation node-by-node, conditioning each prediction on the partial tour. 
% Our non-autoregressive approach is not designed for this setting as it requires two steps: 
% The first step predicts the probability map between all pairs of nodes, independently of other edge predictions. The second step performs greedy search to convert it into a valid tour, adding additional time overhead.
In contrast, our model predicts an edge between any pair of nodes independent of other edge predictions.
The two-step process of getting the predictions from our model and performing greedy search to convert them into a valid tour adds time overhead.

%However, it is unfair to compare our non-autoregressive model with autoregressive approaches in the greedy setting, especially in terms of evaluation time.
%Autoregressive models directly output the TSP tour permutation node-by-node, conditioning each prediction on the partial tour.
%In contrast, our model's prediction of an edge between any pair of nodes is independent of other edge predictions.
%Moreover, the two-step process of getting the output from our model and performing greedy search to convert it into a valid tour adds additional time overhead.

\paragraph{Search/sampling setting}
In general, all learning-based approaches are able to improve performance over the greedy setting by searching or sampling for solutions.
Our graph ConvNet model with beam search outperforms \cite{kool2018attention} in terms of both closeness to optimality and evaluation time when searching/sampling 1,280 solutions.
We attribute our gains in performance to better representation learning for the input graphs through our use of deep architectures with up to 30 graph convolution layers.
In contrast, \cite{kool2018attention} use only 3 graph attention layers.
Despite larger models being more computationally expensive, our graph ConvNet and beam search implementations are highly parallelized for GPU computation, leading to significantly faster evaluation compared to sampling from a reinforcement learning policy.

\paragraph{Combining learned and traditional heuristics}
As seen from the results of \cite{deudon2018learning}, autoregressive models may not produce a local optimum and performance can improve by using a \textit{hybrid} approach of a learned algorithm with a local search heuristic such as 2-OPT. 
The same observation holds for our non-autoregressive model. 
Adding the shortest tour heuristic to beam search boosts our performance at the cost of evaluation time.
Future work shall explore further trade-offs between performance and computation when incorporating heuristics such as 2-OPT into beam search decoding.

\begin{figure}[b!]
    \centering
    \subfloat[TSP20]{
    \includegraphics[width=0.33\textwidth]{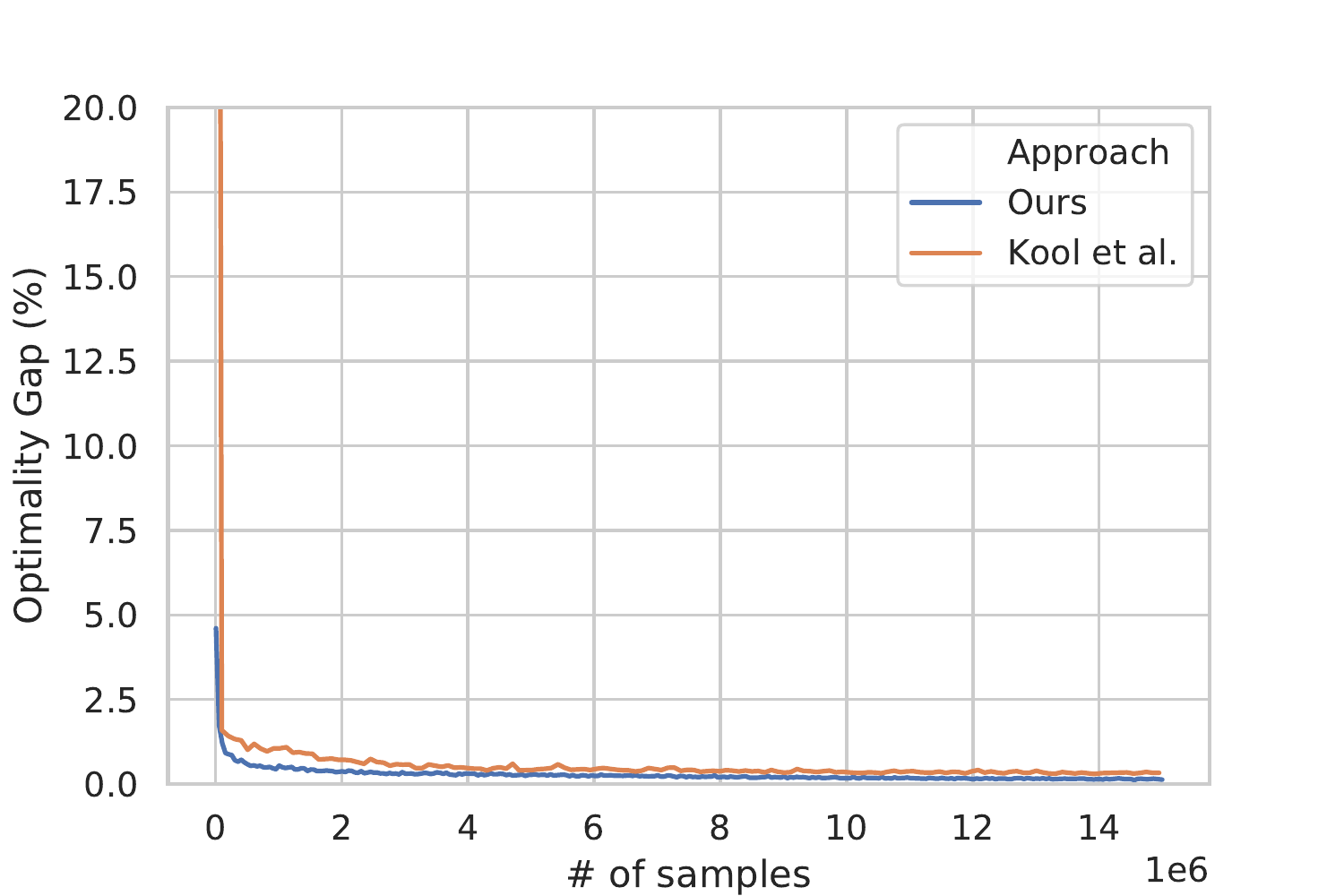}
    \label{fig:val-opt-gap-20}
    }
    \subfloat[TSP50]{
    \includegraphics[width=0.33\textwidth]{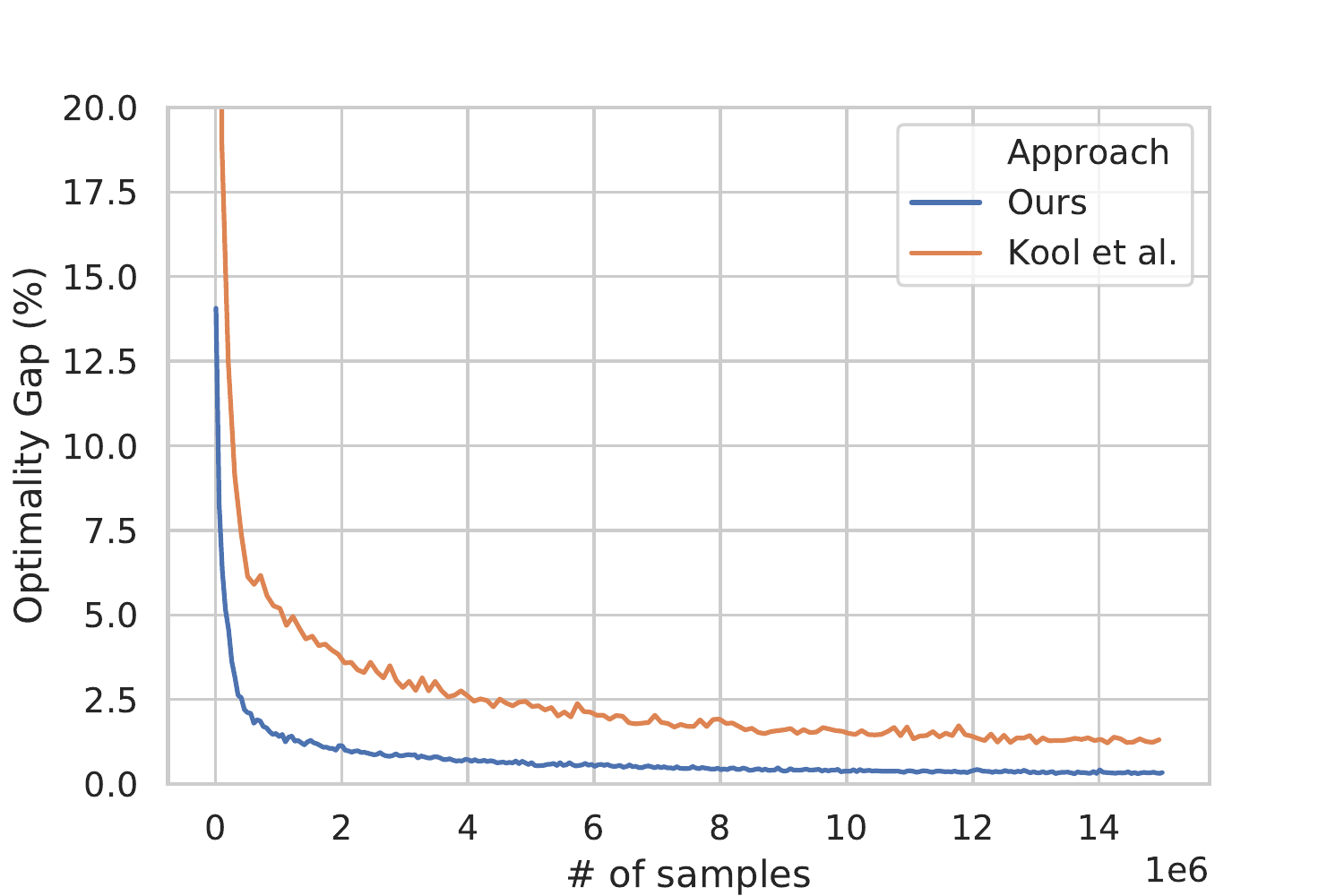}
    \label{fig:val-opt-gap-50}
    }
    \subfloat[TSP100]{
    \includegraphics[width=0.33\textwidth]{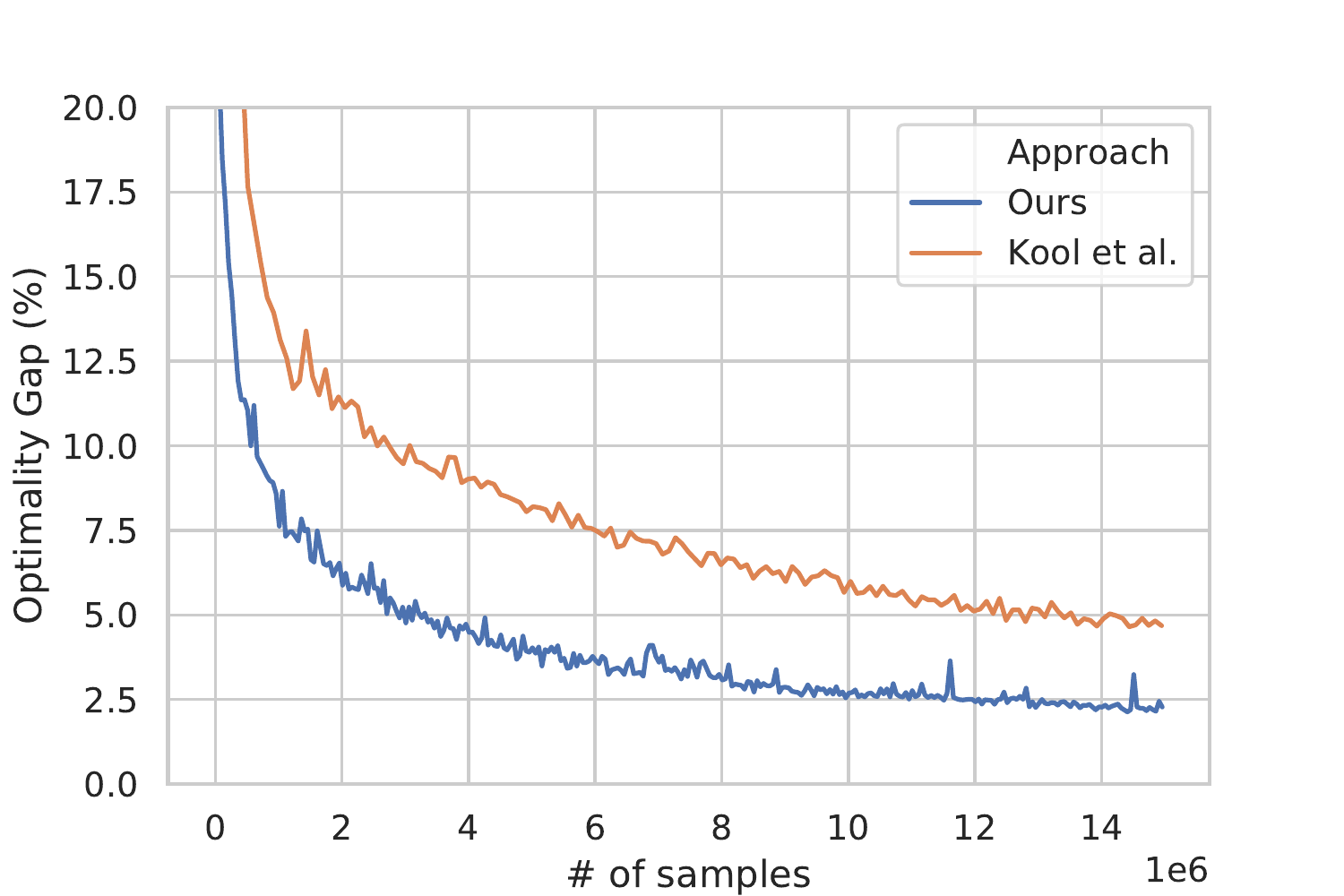}
    \label{fig:val-opt-gap-100}
    }
    \caption{Validation optimality gap vs. Number of training samples for our approach (using beam search with beam width 1,280) and \cite{kool2018attention} (sampling 1,280 solutions).}
    \label{fig:val-opt-gap}
\end{figure}

\paragraph{Sample efficiency for SL vs. RL}
Figure \ref{fig:val-opt-gap} displays the validation optimality gap vs. number of training samples for our approach compared to \cite{kool2018attention}.
For smaller instances (TSP20), both approaches find solutions within 1\% of optimal after seeing less than 500,000 samples.
More training samples are required to solve larger problem instances (TSP50 and TSP100). 
Our supervised training setup is more sample efficient compared to reinforcement learning as we train the model with complete information about the problem whereas RL training is guided by a sparse reward function. 

It is important to note that each training graph is generated on the fly and is unique for RL.
In contrast, our supervised approach randomly selects and repeats training graphs (and their groundtruth solutions) from a fixed set of one million instances.

\paragraph{Generalization to variable problem sizes}

Generalization across TSP instances of various sizes is a highly desirable property for learning combinatorial problems.
Size invariant generalization would allow us to scale up to very large TSP instances while training efficiently on smaller instances.  
In theory, since our model's parameters $\Theta$ are independent of the size of an instance $n$, we can use a model trained on smaller graphs to solve arbitrarily large instances. 

Table \ref{table:generalization} presents the generalization performance of our approach and that of \cite{kool2018attention} by evaluating the best performing models for fixed problem sizes on all other sizes.
We observe drastic drops in performance for our non-autoregressive models, indicating very poor generalization capabilities.
The representations learnt by the graph ConvNet are memorizing patterns for specific graph sizes and are unable to transfer to new graphs.
In contrast, the autoregressive approach of
\cite{kool2018attention} displays a less drastic drop in generalization performance.
In future work, we shall further explore generalization and transfer learning for large-scale combinatorial problems. 

Training logs and additional results on model architecture are presented in Appendix \ref{additional-results}.
Appendix \ref{sl-vs-rl} contains further discussion on supervised learning vs. reinforcement learning for combinatorial problems.
Visualizations and qualitative analysis of solutions produced by our approach are available in Appendix \ref{viz}.

\begin{table}[tb!]
\centering
\caption{Performance of our technique (using beam search with beam width 1,280) and \cite{kool2018attention} (sampling 1,280 solutions) for generalization to variable problem sizes.
The optimality gap is computed w.r.t Concorde. 
}
\label{table:generalization}
\resizebox{\textwidth}{!}{%
\begin{tabular}{l|ccc|ccc|ccc}
\toprule
\multirow{2}{*}{Method/Model} & \multicolumn{3}{c|}{TSP20} & \multicolumn{3}{c|}{TSP50} & \multicolumn{3}{c}{TSP100} \\
 & Tour Len. & Opt. Gap. & Time & Tour Len. & Opt. Gap. & Time & Tour Len. & Opt. Gap. & Time \\
\midrule
Concorde &  $3.84$ & $0.00 \%$ & (1m) & $5.70$ & $0.00 \%$ & (2m) & $7.76$ & $0.00 \%$ & (3m) \\
\midrule
TSP20 Model \citep{kool2018attention} & $3.84$ & $0.08\%$ & (5m) & $5.79$ & $1.78\%$ & (24m) & $9.50$ & $22.61\%$ & (1h) \\
TSP50 Model \citep{kool2018attention} & $3.84$ & $0.35\%$ & (5m) & $5.73$ & $0.52\%$ & (24m) & $7.98$ & $2.95\%$ & (1h) \\
TSP100 Model \citep{kool2018attention} & $3.97$ & $3.78\%$ & (5m) & $5.82$ & $2.33\%$ & (24m) & $7.94$ & $2.26\%$ & (1h) \\
\midrule
TSP20 Model (Ours) & $3.84$ & $0.10\%$ & (20s) & $7.66$ & $34.46\%$ & (2m) & $13.18$ & $69.95\%$ & (10m) \\
TSP50 Model (Ours) & $5.31$ & $38.46\%$ & (20s) & $5.71$ & $0.26\%$ & (2m) & $12.83$ & $65.39\%$ & (10m) \\
TSP100 Model (Ours) & $4.94$ & $28.68\%$ & (20s) & $7.43$ & $30.49\%$ & (2m) & $7.92$ & $2.11\%$ & (10m) \\
\bottomrule
\end{tabular}%
}
\end{table}

\section{Conclusions}
\label{conclusion}

We introduce a novel learning-based approach for
approximately solving the 2D Euclidean Travelling Salesman Problem
using graph ConvNets and beam search. 
For fixed graph sizes, our framework outperforms all previous deep learning techniques in 
terms of solution quality, inference speed and sample efficiency due to better graph representation capacity, highly parallelized implementation and learning from optimal solutions.
Future work shall explore incorporating transfer learning and reinforcement learning into our framework in order to generalize to large-scale problem instances and tackle previously un-encountered combinatorial problems beyond TSP.

\section*{Acknowledgement}
The authors thank Victor Getty for helpful comments and discussions.
XB is supported in part by NRF Fellowship NRFF2017-10.

\bibliographystyle{plainnat}
\bibliography{main}

\begin{thebibliography}{43}
\providecommand{\natexlab}[1]{#1}
\providecommand{\url}[1]{\texttt{#1}}
\expandafter\ifx\csname urlstyle\endcsname\relax
  \providecommand{\doi}[1]{doi: #1}\else
  \providecommand{\doi}{doi: \begingroup \urlstyle{rm}\Url}\fi

\bibitem[Angeniol et~al.(1988)Angeniol, Vaubois, and
  Le~Texier]{angeniol1988self}
Bernard Angeniol, Gael De La~Croix Vaubois, and Jean-Yves Le~Texier.
\newblock Self-organizing feature maps and the travelling salesman problem.
\newblock \emph{Neural Networks}, 1\penalty0 (4):\penalty0 289--293, 1988.

\bibitem[Applegate et~al.(2003)Applegate, Bixby, Chv{\'a}tal, and
  Cook]{applegate2003implementing}
David Applegate, Robert Bixby, Va{\v{s}}ek Chv{\'a}tal, and William Cook.
\newblock Implementing the dantzig-fulkerson-johnson algorithm for large
  traveling salesman problems.
\newblock \emph{Mathematical programming}, 97\penalty0 (1-2):\penalty0 91--153,
  2003.

\bibitem[Applegate et~al.(2006)Applegate, Bixby, Chvatal, and
  Cook]{applegate2006traveling}
David~L Applegate, Robert~E Bixby, Vasek Chvatal, and William~J Cook.
\newblock \emph{The traveling salesman problem: a computational study}.
\newblock Princeton university press, 2006.

\bibitem[Bahdanau et~al.(2014)Bahdanau, Cho, and Bengio]{bahdanau2014neural}
Dzmitry Bahdanau, Kyunghyun Cho, and Yoshua Bengio.
\newblock Neural machine translation by jointly learning to align and
  translate.
\newblock \emph{arXiv preprint arXiv:1409.0473}, 2014.

\bibitem[Bello et~al.(2016)Bello, Pham, Le, Norouzi, and
  Bengio]{bello2016neural}
Irwan Bello, Hieu Pham, Quoc~V Le, Mohammad Norouzi, and Samy Bengio.
\newblock Neural combinatorial optimization with reinforcement learning.
\newblock \emph{arXiv preprint arXiv:1611.09940}, 2016.

\bibitem[Bengio et~al.(2018)Bengio, Lodi, and Prouvost]{bengio2018machine}
Yoshua Bengio, Andrea Lodi, and Antoine Prouvost.
\newblock Machine learning for combinatorial optimization: a methodological
  tour d'horizon.
\newblock \emph{arXiv preprint arXiv:1811.06128}, 2018.

\bibitem[Bresson and Laurent(2017)]{bresson2017residual}
Xavier Bresson and Thomas Laurent.
\newblock Residual gated graph convnets.
\newblock \emph{arXiv preprint arXiv:1711.07553}, 2017.

\bibitem[Bronstein et~al.(2017)Bronstein, Bruna, LeCun, Szlam, and
  Vandergheynst]{bronstein2017geometric}
Michael~M Bronstein, Joan Bruna, Yann LeCun, Arthur Szlam, and Pierre
  Vandergheynst.
\newblock Geometric deep learning: going beyond euclidean data.
\newblock \emph{IEEE Signal Processing Magazine}, 34\penalty0 (4):\penalty0
  18--42, 2017.

\bibitem[Bruna et~al.(2013)Bruna, Zaremba, Szlam, and LeCun]{bruna2013spectral}
Joan Bruna, Wojciech Zaremba, Arthur Szlam, and Yann LeCun.
\newblock Spectral networks and locally connected networks on graphs.
\newblock \emph{arXiv preprint arXiv:1312.6203}, 2013.

\bibitem[Croes(1958)]{croes1958method}
Georges~A Croes.
\newblock A method for solving traveling-salesman problems.
\newblock \emph{Operations research}, 6\penalty0 (6):\penalty0 791--812, 1958.

\bibitem[Dai et~al.(2016)Dai, Dai, and Song]{dai2016discriminative}
Hanjun Dai, Bo~Dai, and Le~Song.
\newblock Discriminative embeddings of latent variable models for structured
  data.
\newblock In \emph{International conference on machine learning}, pages
  2702--2711, 2016.

\bibitem[Dai et~al.(2017)Dai, Khalil, Zhang, Dilkina, and
  Song]{dai2017learning}
Hanjun Dai, Elias Khalil, Yuyu Zhang, Bistra Dilkina, and Le~Song.
\newblock Learning combinatorial optimization algorithms over graphs.
\newblock In \emph{Advances in Neural Information Processing Systems}, pages
  6348--6358, 2017.

\bibitem[Dantzig et~al.(1954)Dantzig, Fulkerson, and
  Johnson]{dantzig1954solution}
George Dantzig, Ray Fulkerson, and Selmer Johnson.
\newblock Solution of a large-scale traveling-salesman problem.
\newblock \emph{Journal of the operations research society of America},
  2\penalty0 (4):\penalty0 393--410, 1954.

\bibitem[Defferrard et~al.(2016)Defferrard, Bresson, and
  Vandergheynst]{defferrard2016convolutional}
Micha{\"e}l Defferrard, Xavier Bresson, and Pierre Vandergheynst.
\newblock Convolutional neural networks on graphs with fast localized spectral
  filtering.
\newblock In \emph{Advances in neural information processing systems}, pages
  3844--3852, 2016.

\bibitem[Deudon et~al.(2018)Deudon, Cournut, Lacoste, Adulyasak, and
  Rousseau]{deudon2018learning}
Michel Deudon, Pierre Cournut, Alexandre Lacoste, Yossiri Adulyasak, and
  Louis-Martin Rousseau.
\newblock Learning heuristics for the tsp by policy gradient.
\newblock In \emph{International Conference on the Integration of Constraint
  Programming, Artificial Intelligence, and Operations Research}, pages
  170--181. Springer, 2018.

\bibitem[Fort(1988)]{fort1988solving}
JC~Fort.
\newblock Solving a combinatorial problem via self-organizing process: An
  application of the kohonen algorithm to the traveling salesman problem.
\newblock \emph{Biological cybernetics}, 59\penalty0 (1):\penalty0 33--40,
  1988.

\bibitem[Hamilton et~al.(2017)Hamilton, Ying, and
  Leskovec]{hamilton2017inductive}
Will Hamilton, Zhitao Ying, and Jure Leskovec.
\newblock Inductive representation learning on large graphs.
\newblock In \emph{Advances in Neural Information Processing Systems}, pages
  1024--1034, 2017.

\bibitem[He et~al.(2016)He, Zhang, Ren, and Sun]{he2016deep}
Kaiming He, Xiangyu Zhang, Shaoqing Ren, and Jian Sun.
\newblock Deep residual learning for image recognition.
\newblock In \emph{Proceedings of the IEEE conference on computer vision and
  pattern recognition}, pages 770--778, 2016.

\bibitem[Hopfield and Tank(1985)]{hopfield1985neural}
John~J Hopfield and David~W Tank.
\newblock “neural” computation of decisions in optimization problems.
\newblock \emph{Biological cybernetics}, 52\penalty0 (3):\penalty0 141--152,
  1985.

\bibitem[Ioffe and Szegedy(2015)]{ioffe2015batch}
Sergey Ioffe and Christian Szegedy.
\newblock Batch normalization: Accelerating deep network training by reducing
  internal covariate shift.
\newblock \emph{arXiv preprint arXiv:1502.03167}, 2015.

\bibitem[Kaempfer and Wolf(2018)]{kaempfer2018learning}
Yoav Kaempfer and Lior Wolf.
\newblock Learning the multiple traveling salesmen problem with permutation
  invariant pooling networks.
\newblock \emph{arXiv preprint arXiv:1803.09621}, 2018.

\bibitem[Kingma and Ba(2014)]{kingma2014adam}
Diederik~P Kingma and Jimmy Ba.
\newblock Adam: A method for stochastic optimization.
\newblock \emph{arXiv preprint arXiv:1412.6980}, 2014.

\bibitem[Kipf and Welling(2016)]{kipf2016semi}
Thomas~N Kipf and Max Welling.
\newblock Semi-supervised classification with graph convolutional networks.
\newblock \emph{arXiv preprint arXiv:1609.02907}, 2016.

\bibitem[Kool et~al.(2019)Kool, van Hoof, and Welling]{kool2018attention}
Wouter Kool, Herke van Hoof, and Max Welling.
\newblock Attention, learn to solve routing problems!
\newblock In \emph{International Conference on Learning Representations}, 2019.
\newblock URL \url{https://openreview.net/forum?id=ByxBFsRqYm}.

\bibitem[La~Maire and Mladenov(2012)]{la2012comparison}
Bert~FJ La~Maire and Valeri~M Mladenov.
\newblock Comparison of neural networks for solving the travelling salesman
  problem.
\newblock In \emph{11th Symposium on Neural Network Applications in Electrical
  Engineering}, pages 21--24. IEEE, 2012.

\bibitem[Li et~al.(2018)Li, Chen, and Koltun]{li2018combinatorial}
Zhuwen Li, Qifeng Chen, and Vladlen Koltun.
\newblock Combinatorial optimization with graph convolutional networks and
  guided tree search.
\newblock In \emph{Advances in Neural Information Processing Systems}, pages
  539--548, 2018.

\bibitem[Marcheggiani and Titov(2017)]{marcheggiani2017encoding}
Diego Marcheggiani and Ivan Titov.
\newblock Encoding sentences with graph convolutional networks for semantic
  role labeling.
\newblock \emph{arXiv preprint arXiv:1703.04826}, 2017.

\bibitem[Medress et~al.(1977)Medress, Cooper, Forgie, Green, Klatt, O'Malley,
  Neuburg, Newell, Reddy, Ritea, et~al.]{medress1977speech}
Mark~F. Medress, Franklin~S Cooper, Jim~W. Forgie, CC~Green, Dennis~H. Klatt,
  Michael~H. O'Malley, Edward~P Neuburg, Allen Newell, DR~Reddy, B~Ritea,
  et~al.
\newblock Speech understanding systems: Report of a steering committee.
\newblock \emph{Artificial Intelligence}, 9\penalty0 (3):\penalty0 307--316,
  1977.

\bibitem[Mittal et~al.(2019)Mittal, Dhawan, Medya, Ranu, and
  Singh]{mittal2019learning}
Akash Mittal, Anuj Dhawan, Sourav Medya, Sayan Ranu, and Ambuj Singh.
\newblock Learning heuristics over large graphs via deep reinforcement
  learning.
\newblock \emph{arXiv preprint arXiv:1903.03332}, 2019.

\bibitem[Mnih et~al.(2013)Mnih, Kavukcuoglu, Silver, Graves, Antonoglou,
  Wierstra, and Riedmiller]{mnih2013playing}
Volodymyr Mnih, Koray Kavukcuoglu, David Silver, Alex Graves, Ioannis
  Antonoglou, Daan Wierstra, and Martin Riedmiller.
\newblock Playing atari with deep reinforcement learning.
\newblock \emph{arXiv preprint arXiv:1312.5602}, 2013.

\bibitem[Nazari et~al.(2018)Nazari, Oroojlooy, Snyder, and
  Tak{\'a}c]{nazari2018reinforcement}
Mohammadreza Nazari, Afshin Oroojlooy, Lawrence Snyder, and Martin Tak{\'a}c.
\newblock Reinforcement learning for solving the vehicle routing problem.
\newblock In \emph{Advances in Neural Information Processing Systems}, pages
  9861--9871, 2018.

\bibitem[Nowak et~al.(2017)Nowak, Villar, Bandeira, and Bruna]{nowak2017note}
Alex Nowak, Soledad Villar, Afonso~S Bandeira, and Joan Bruna.
\newblock A note on learning algorithms for quadratic assignment with graph
  neural networks.
\newblock \emph{arXiv preprint arXiv:1706.07450}, 2017.

\bibitem[Padberg and Rinaldi(1991)]{padberg1991branch}
Manfred Padberg and Giovanni Rinaldi.
\newblock A branch-and-cut algorithm for the resolution of large-scale
  symmetric traveling salesman problems.
\newblock \emph{SIAM review}, 33\penalty0 (1):\penalty0 60--100, 1991.

\bibitem[Papadimitriou(1977)]{papadimitriou1977euclidean}
Christos~H Papadimitriou.
\newblock The euclidean travelling salesman problem is np-complete.
\newblock \emph{Theoretical computer science}, 4\penalty0 (3):\penalty0
  237--244, 1977.

\bibitem[Scarselli et~al.(2009)Scarselli, Gori, Tsoi, Hagenbuchner, and
  Monfardini]{scarselli2009graph}
Franco Scarselli, Marco Gori, Ah~Chung Tsoi, Markus Hagenbuchner, and Gabriele
  Monfardini.
\newblock The graph neural network model.
\newblock \emph{IEEE Transactions on Neural Networks}, 20\penalty0
  (1):\penalty0 61--80, 2009.

\bibitem[Smith(1999)]{smith1999neural}
Kate~A Smith.
\newblock Neural networks for combinatorial optimization: a review of more than
  a decade of research.
\newblock \emph{INFORMS Journal on Computing}, 11\penalty0 (1):\penalty0
  15--34, 1999.

\bibitem[Sukhbaatar et~al.(2016)Sukhbaatar, Szlam, and
  Fergus]{sukhbaatar2016gcn}
Sainbayar Sukhbaatar, Arthur Szlam, and Rob Fergus.
\newblock Learning multiagent communication with backpropagation.
\newblock In \emph{Advances in Neural Information Processing Systems}, pages
  2244--2252, 2016.
\newblock URL \url{https://openreview.net/forum?id=ByxBFsRqYm}.

\bibitem[Sutskever et~al.(2014)Sutskever, Vinyals, and
  Le]{sutskever2014sequence}
Ilya Sutskever, Oriol Vinyals, and Quoc~VV Le.
\newblock Sequence to sequence learning with neural networks.
\newblock In \emph{Advances in Neural Information Processing Systems}, pages
  3104--3112, 2014.

\bibitem[Veli{\v{c}}kovi{\'c} et~al.(2017)Veli{\v{c}}kovi{\'c}, Cucurull,
  Casanova, Romero, Lio, and Bengio]{velivckovic2017graph}
Petar Veli{\v{c}}kovi{\'c}, Guillem Cucurull, Arantxa Casanova, Adriana Romero,
  Pietro Lio, and Yoshua Bengio.
\newblock Graph attention networks.
\newblock \emph{arXiv preprint arXiv:1710.10903}, 2017.

\bibitem[Venkatakrishnan et~al.(2018)Venkatakrishnan, Alizadeh, and
  Viswanath]{venkatakrishnan2018graph2seq}
Shaileshh~Bojja Venkatakrishnan, Mohammad Alizadeh, and Pramod Viswanath.
\newblock Graph2seq: Scalable learning dynamics for graphs.
\newblock \emph{arXiv preprint arXiv:1802.04948}, 2018.

\bibitem[Vinyals et~al.(2015)Vinyals, Fortunato, and
  Jaitly]{vinyals2015pointer}
Oriol Vinyals, Meire Fortunato, and Navdeep Jaitly.
\newblock Pointer networks.
\newblock In \emph{Advances in Neural Information Processing Systems}, pages
  2692--2700, 2015.

\bibitem[Williams(1992)]{williams1992simple}
Ronald~J Williams.
\newblock Simple statistical gradient-following algorithms for connectionist
  reinforcement learning.
\newblock \emph{Machine learning}, 8\penalty0 (3-4):\penalty0 229--256, 1992.

\bibitem[Wu et~al.(2016)Wu, Schuster, Chen, Le, Norouzi, Macherey, Krikun, Cao,
  Gao, Macherey, et~al.]{wu2016google}
Yonghui Wu, Mike Schuster, Zhifeng Chen, Quoc~V Le, Mohammad Norouzi, Wolfgang
  Macherey, Maxim Krikun, Yuan Cao, Qin Gao, Klaus Macherey, et~al.
\newblock Google's neural machine translation system: Bridging the gap between
  human and machine translation.
\newblock \emph{arXiv preprint arXiv:1609.08144}, 2016.

\end{thebibliography}

\newpage
\appendix
\section{Summary Statistics for Datasets}
\label{dataset-stats}

Summary statistics for various TSP datasets are presented in Table \ref{table:datasets}.
We include TSP10 and TSP30 in addition to TSP20, TSP50 and TSP100.
The approximate solver timings for Concorde are computed for a single-thread program on a 32-core CPU server under average load. 
Naturally, Concorde requires longer durations to find exact solutions as problem size increases. 
Generating datasets for problem sizes larger than 1,000 nodes would be impractical on a similar machine. 
% (Generating 1 Million instances of TSP100 took roughly four days of computation.)

\begin{table}[h!]
\centering
\caption{Summary statistics for TSP datasets. \emph{Solver Time} is the average time taken by Concorde to solve each instance on a 32-core CPU server under average load. Each \emph{Tour Len.} column denotes the average TSP tour length over the corresponding set, and the \emph{Std. dev.} columns denote the standard deviation of the same.}
\label{table:datasets}
\resizebox{\textwidth}{!}{%
\begin{tabular}{lccccccc}
\toprule
\multirow{2}{*}{Problem} & Solver Time & \multicolumn{2}{c}{Training set} & \multicolumn{2}{c}{Validation set} & \multicolumn{2}{c}{Test set} \\
 & (approx.)  & Tour Len. & Std. dev. & Tour Len. & Std. dev. & Tour Len. & Std. dev. \\
 \midrule
TSP10 & $1$ms & $2.869$ & $0.337$ & $2.870$ & $0.337$ & $2.871$ & $0.336$ \\
TSP20 & $1$ms & $3.829$ & $0.304$ & $3.830$ & $0.306$ & $3.830$ & $0.303$ \\
TSP30 & $1$ms & $4.555$ & $0.281$ & $4.540$ & $0.280$ & $4.539$ & $0.283$ \\
TSP50 & $10$ms & $5.693$ & $0.252$ & $5.693$ & $0.252$ & $5.692$ & $0.252$  \\
TSP100 & $250$ms & $7.766$ & $0.230$ & $7.765$ & $0.231$ & $7.764$ & $0.228$ \\
\bottomrule
\end{tabular}%
}
\end{table}

\section{Graph Convolution Layer Details} 
\label{model-details}

For the graph ConvNet described in Section \ref{gcn}, 
let  $x_i^{\ell}$ denote the feature vector at layer $\ell$ associated with node $i$. 
The activation $x_i^{\ell+1}$ at the next layer is obtained by applying a non-linear transformation to the feature vectors $x_{j}^{\ell}$ for all nodes $j$ in the neighborhood of node $i$ (defined by the graph structure). Thus, the most generic version of a feature vector $x_i^{\ell+1}$ at vertex $i$ in a graph ConvNet is:
\begin{equation} 
\label{generic-node}
x_{i}^{\ell+1} = f \left( \; x_i^{\ell}  \;   , \;  \{ x_{j}^{\ell}: j \sim i \}  \;  \right)
\end{equation}
where $\{ j \sim i \}$ denotes the set of neighboring nodes centered at node $i$. In other words, a graph ConvNet is defined by a mapping $f$ taking as input a vector  $x_i^{\ell}$ (the feature vector of the center vertex) as well as an un-ordered set of vectors $\{ x_{j}^{\ell}\}$ (the feature vectors of all neighboring vertices). 
The arbitrary choice of the mapping $f$ defines an instantiation of a class of graph neural networks such as \cite{sukhbaatar2016gcn,kipf2016semi,hamilton2017inductive}.

For this work, we leverage the graph ConvNet architecture introduced in \cite{bresson2017residual} by defining node features $x_i^{\ell+1}$ and edge features $e_{ij}^{\ell+1}$ as follows:
\begin{eqnarray}
x_i^{\ell+1} &=& x_i^{\ell} + \text{ReLU} \Big( \text{BN} \Big( W_1^\ell x_i^{\ell} + \sum_{j\sim i} \eta_{ij}^{\ell} \odot W_2^\ell x_j^{\ell} \Big)\Big) \ \textrm{with} \ \eta_{ij}^{\ell} = \frac{\sigma(e_{ij}^{\ell})}{\sum_{j'\sim i} \sigma(e_{ij'}^{\ell}) + \varepsilon }, \label{eqn:gcn-node2}\\
e_{ij}^{\ell+1} &=& e^\ell_{ij} + \text{ReLU}\Big( \text{BN} \Big( W_3^\ell e_{ij}^{\ell} + W_4^\ell x^{\ell}_i + W_5^\ell x^{\ell}_j  \Big)\Big),\label{eqn:gcn-edge2}
\end{eqnarray}
where $W \in \mathbb{R}^{h \times h} $, $\sigma$ is the sigmoid function, $\varepsilon$ is a small value, $\text{ReLU}$ is the rectified linear unit, and $\text{BN}$ stands for batch normalization. 
At the input layer, we have $x_i^{\ell=0}=\alpha_i$ and $e_{ij}^{\ell=0}=\beta_{ij}$. 

Eq. \eqref{eqn:gcn-node2} is similar to a learnable non-diffusion process on graphs where the diffusion time is the number $\ell$ of layers. 
As arbitrary graphs have no specific orientations (up, down, left, right), a diffusion process on graphs is consequently \textit{isotropic}, making all neighbors equally important. 
However, this may not be true in general, e.g. a neighbor in the same community of a node shares different information than a neighbor in a separate community. 
We make the diffusion process \textit{anisotropic} by point-wise multiplication operations with learneable normalized edge gates $e_{ij}^{\ell}$ such as in \citep{marcheggiani2017encoding}. 
Eq. \eqref{eqn:gcn-edge2} also represents a learnable non-diffusion process on graphs for the edge features. The network learns the best edge representations for encoding the information flow on the graph structure. 

\begin{figure}[h!]
    \center{\includegraphics[width=\textwidth]{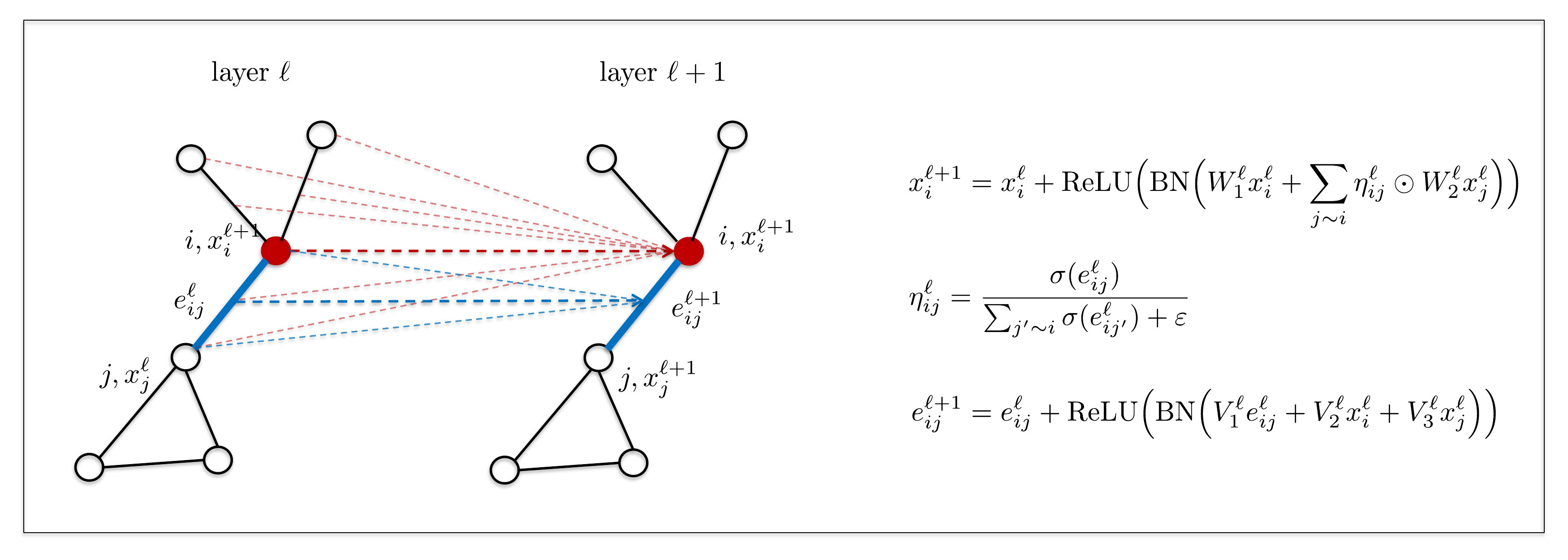}}
    \caption{The proposed graph convolution layer for computing $h$-dimensional representations $x_i$ for each node $i$ and $e_{ij}$ for the edge between each node $i$ and $j$ in the graph. The red and blue arrows respectively represent the node and edge information used to compute their representation at the next layer. Multiple layers of graph convolution are applied to progressively extract more and more compositional features of the input graph.}
    \label{fig:conv}
\end{figure}

Additionally, we perform batch normalization \citep{ioffe2015batch} to enable fast training of deep architectures.
Residual connections in Eq. \eqref{eqn:gcn-node2} are essential to minimize the effect of the vanishing gradient problem on backpropagation \citep{he2016deep}. 
Figure \ref{fig:conv} illustrates the model.

\section{Additional Results}
\label{additional-results}

\begin{figure}[H]
    \centering
    \subfloat[Learning Rates]{
    \includegraphics[width=0.25\textwidth]{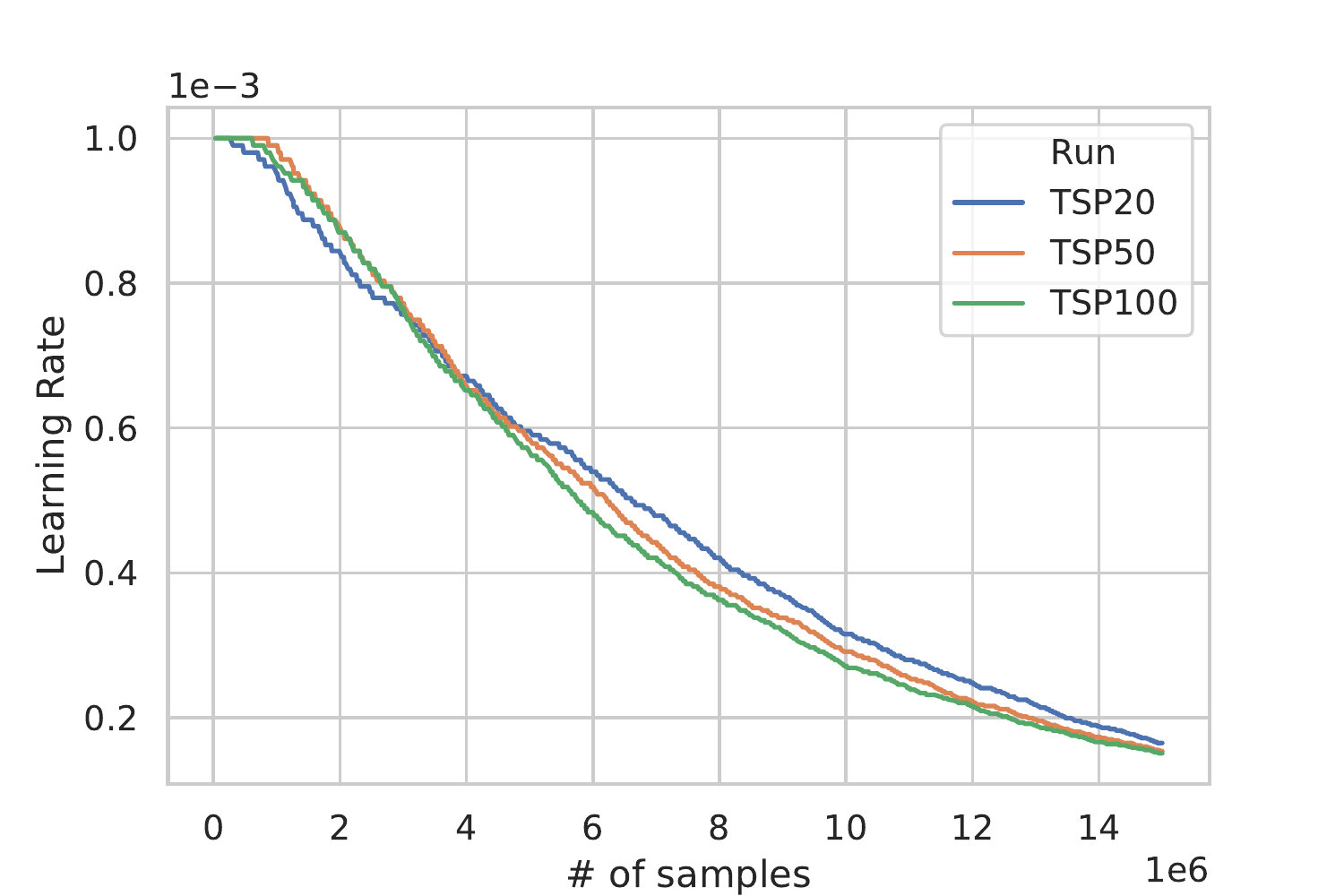}
    \label{fig:lr}
    }
    \subfloat[TSP20]{
    \includegraphics[width=0.25\textwidth]{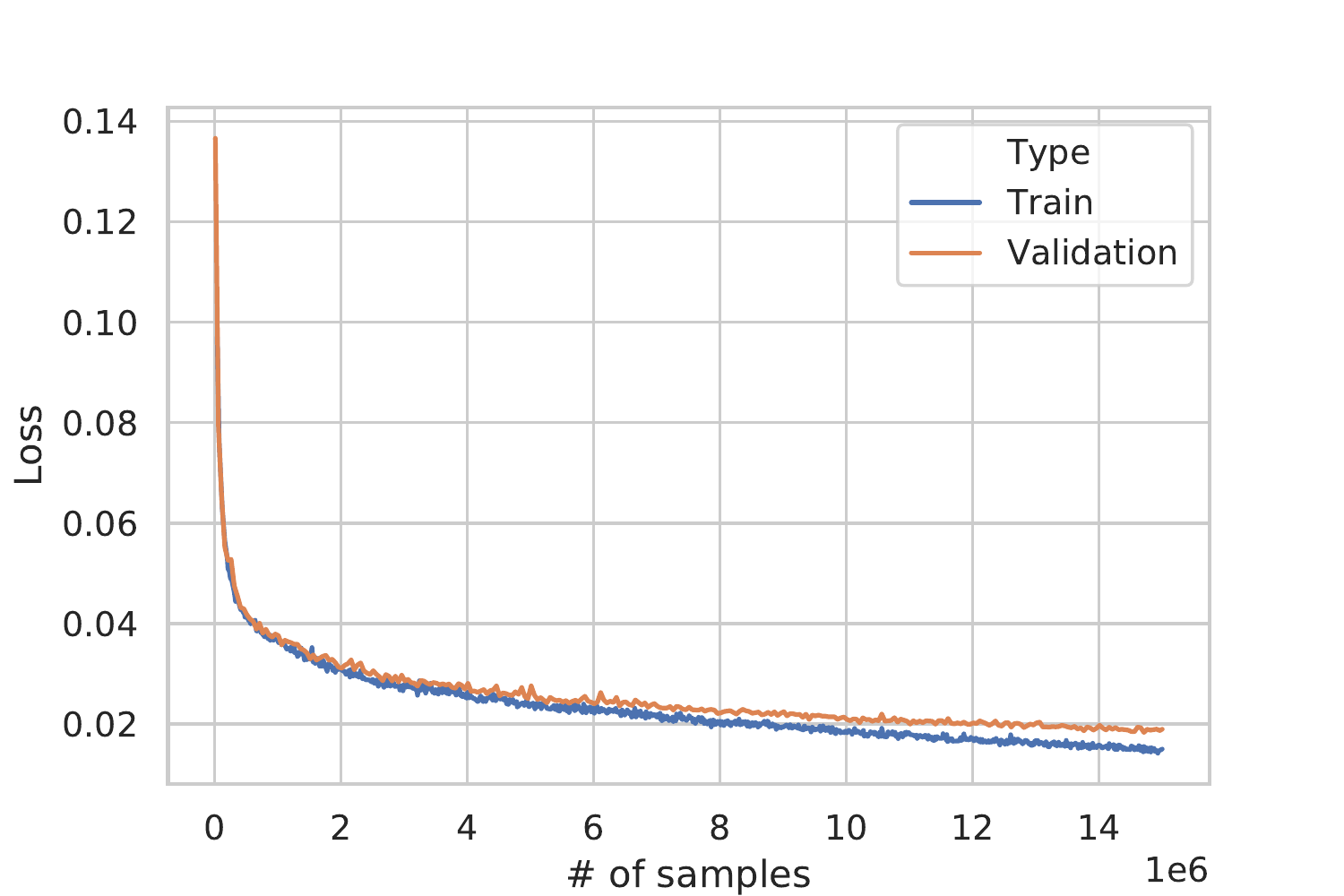}
    \label{fig:loss-20}
    }
    \subfloat[TSP50]{
    \includegraphics[width=0.25\textwidth]{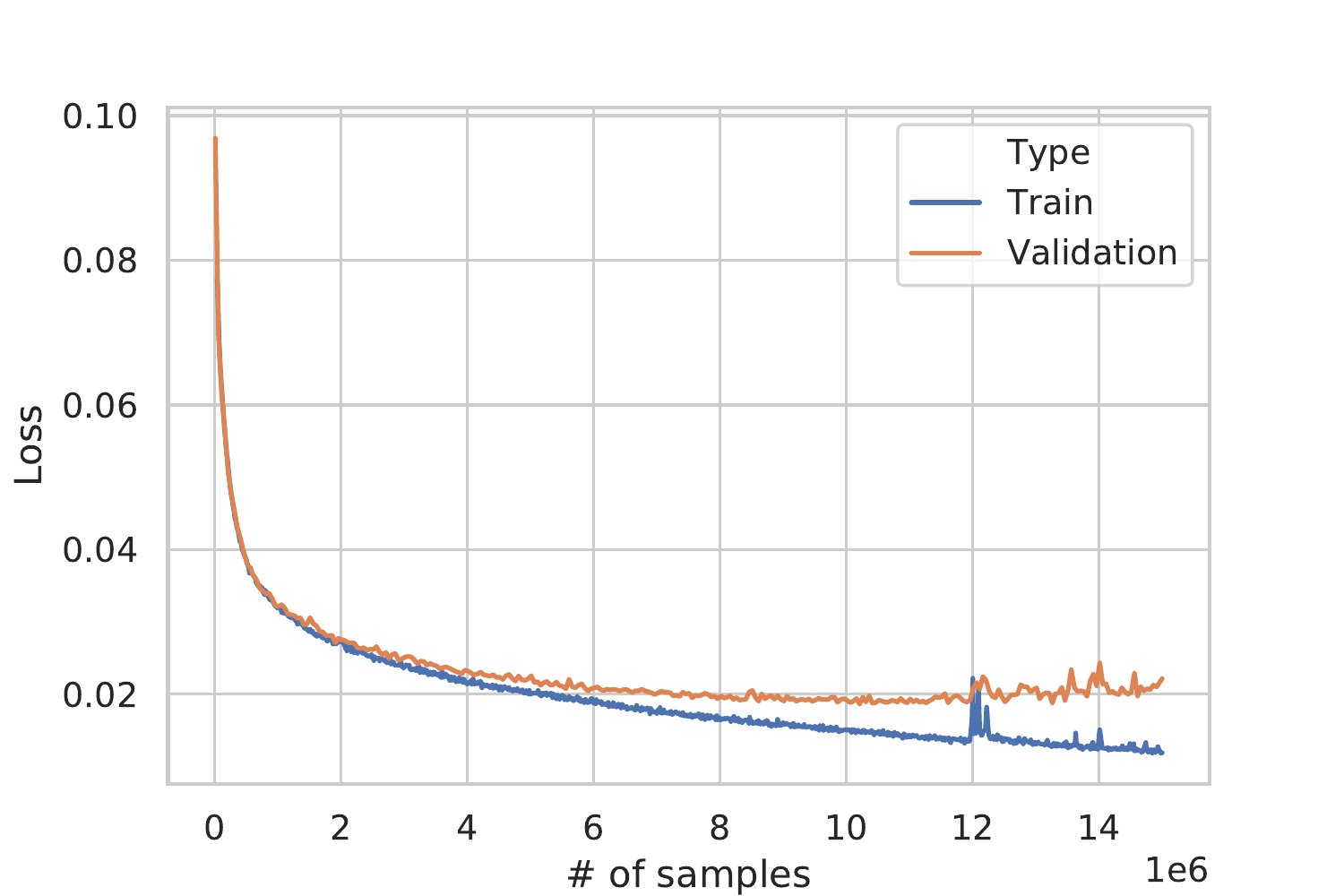}
    \label{fig:loss-50}
    }
    \subfloat[TSP100]{
    \includegraphics[width=0.25\textwidth]{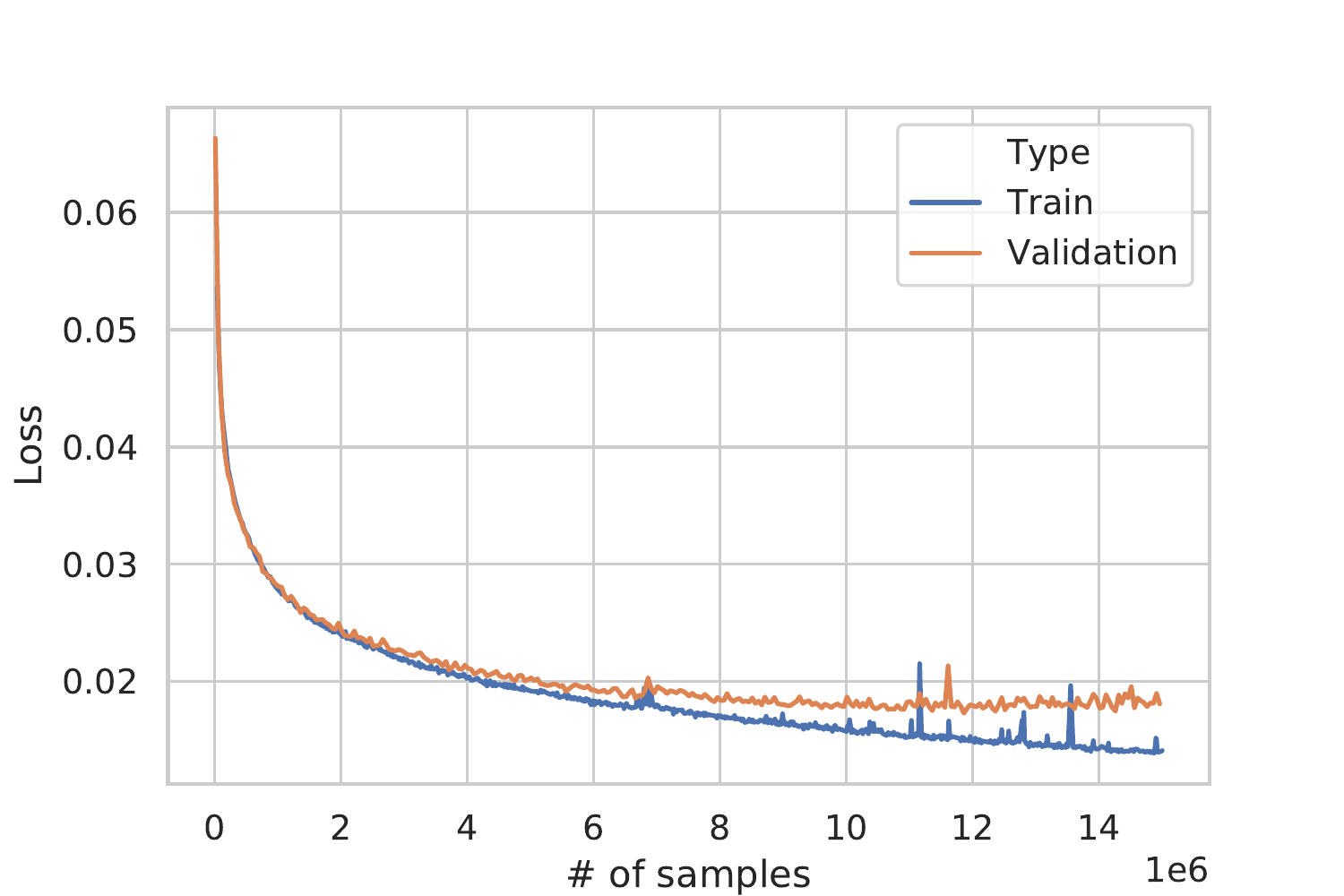}
    \label{fig:loss-100}
    }
    \caption{Learning rate or Training/Validation loss vs. Number of training samples.}
    \label{fig:loss}
\end{figure}

\paragraph{Training logs}
Figure \ref{fig:loss} displays the learning rate and loss values vs. number of training samples for various runs.
Decaying the learning rate to very small values allows the training loss to smoothly decrease.
Loss curves for TSP50 and TSP100 show that models start overfitting to the training set after seeing approximately 4 million samples. 
However, as seen in Figure \ref{fig:val-opt-gap}, the validation optimality gap does not get worse as models overfit to training data.
% This indicates that generalization to unseen problem instances of the same size as those seen in training is not a good indicator of model performance.

For faster training, TSP100 models were trained using four Nvidia 1080Ti GPUs. However, it is not essential to use a multi-GPU setup for training or evaluating our models: the same results can be attained with a single GPU by training longer.

\begin{figure}[H]
    \centering
    \subfloat[Model capacity (TSP50)]{
    \includegraphics[width=0.33\textwidth]{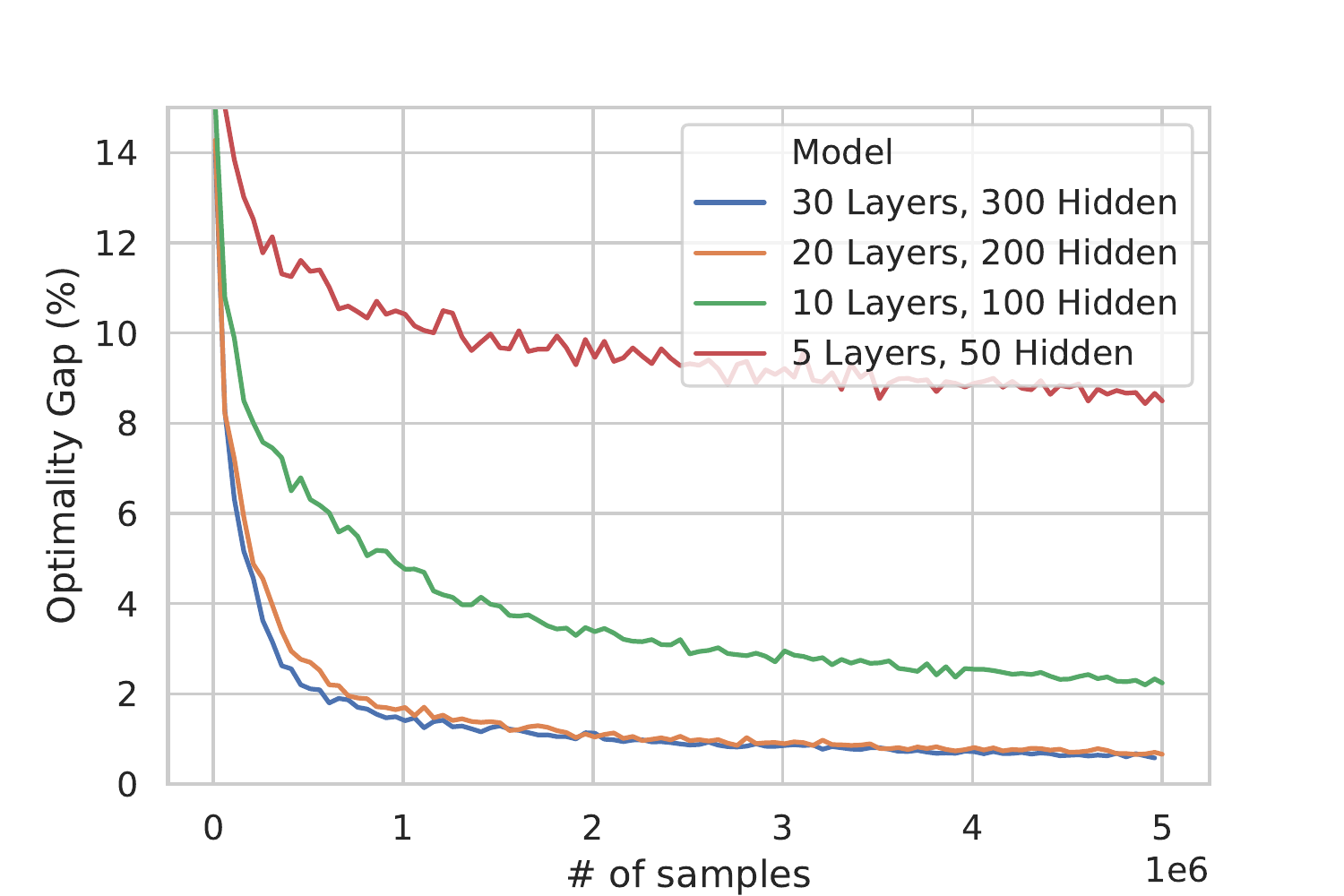}
    \label{fig:val-opt-gap-cap}
    }
    \subfloat[Beam width]{
    \includegraphics[width=0.33\textwidth]{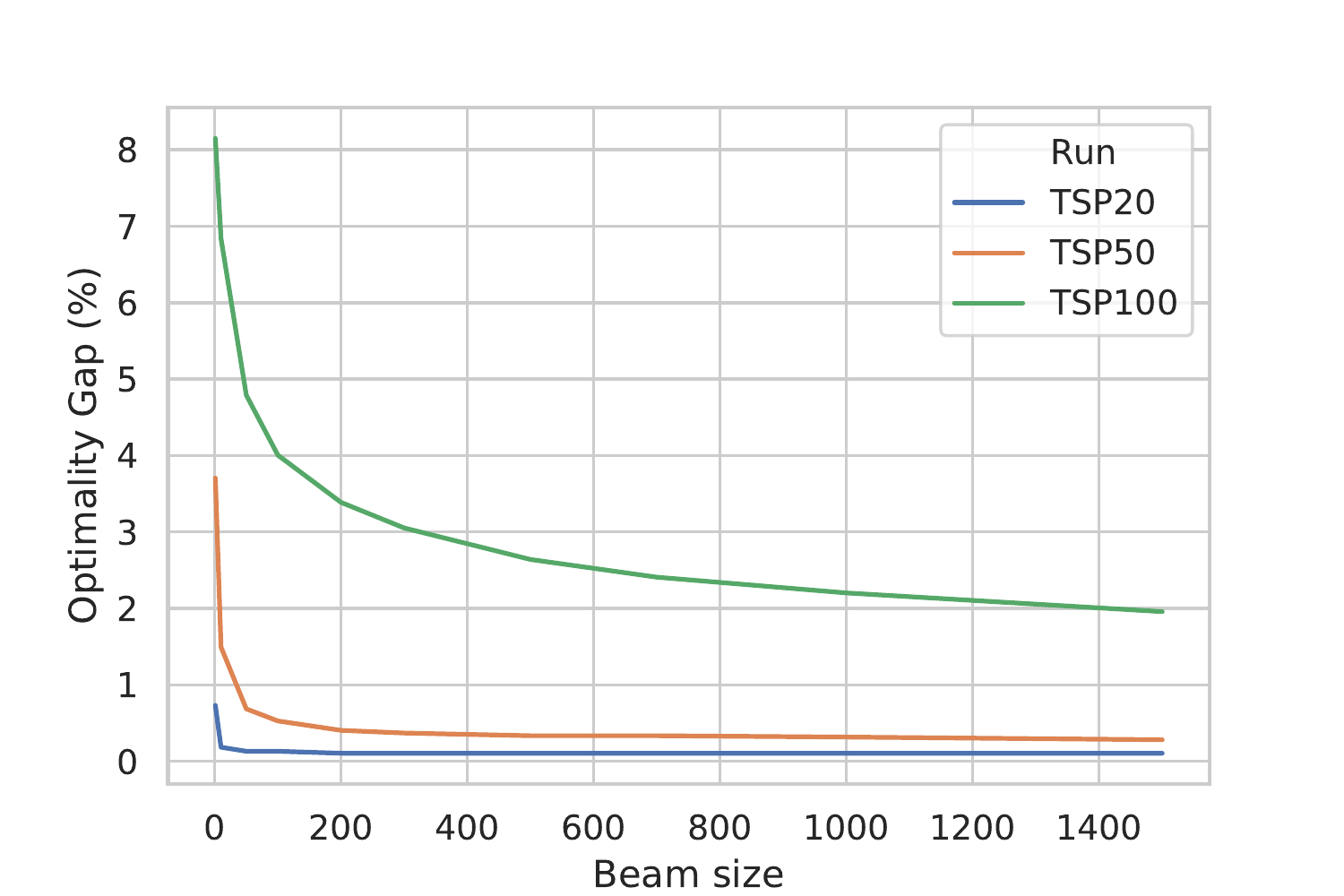}
    \label{fig:val-opt-gap-beam}
    }
    \subfloat[Attention type (TSP50)]{
    \includegraphics[width=0.33\textwidth]{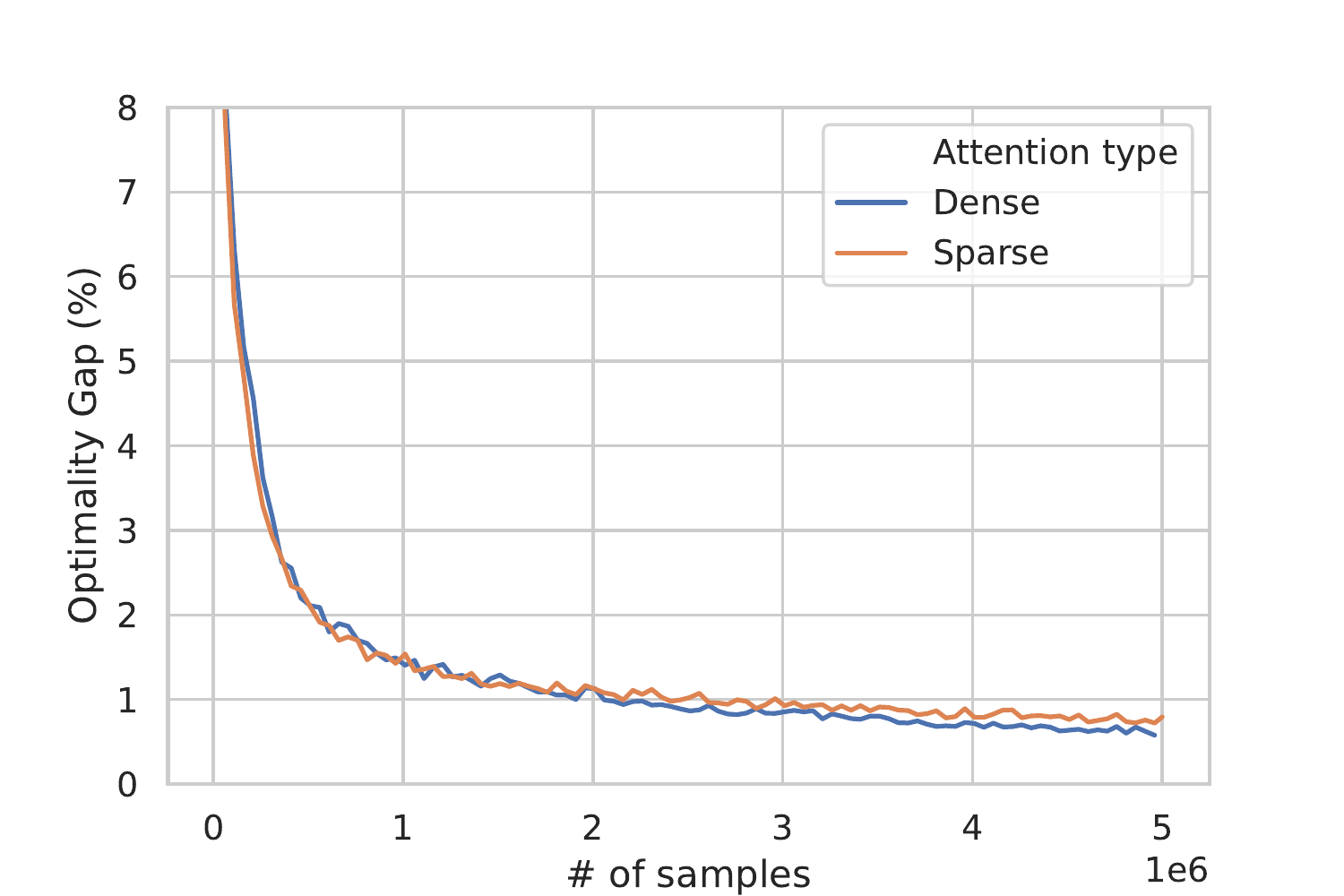}
    \label{fig:val-opt-gap-attn}
    }
    \caption{Impact of model architecture on validation optimality gap.}
    \label{fig:val-opt-gap-extra}
\end{figure}

\paragraph{Model capacity}
Figure \ref{fig:val-opt-gap-cap} presents the validation optimality gap vs. number of training samples of TSP50 for various model capacities (defined as the number of graph convolution layers and hidden dimension).
In general, we found that smaller models are able to learn smaller problem sizes at approximately the same closeness to optimality as larger models.
Increasing model capacity leads to longer training time but is essential for scaling up to large problem sizes.

For consistency of analysis across problem sizes, our main results use models with the maximum capacity possible (30 layers, 300 hidden dimension) for our hardware setup (4 $\times$ Nvidia 1080Ti GPUs) and the largest problem size (TSP100).

\paragraph{Beam width}
Figure \ref{fig:val-opt-gap-beam} presents the validation optimality gap vs. beam width for various problem sizes.
For smaller problem sizes, increasing beam width beyond 200 has a minor impact on performance.
For TSP100, using large beam widths is essential for performance.

Reinforcement learning approaches such as \cite{kool2018attention} sample 1,280 tours from the learnt policy (in < 1s on a single GPU for their model) and report the shortest among them as the final solution. 
For our main results, we use a beam width of 1,280 in order to directly compare with their approach.
Due to the non-autoregressive nature of our approach, we can search using beam widths considerably larger than 1,280 within 1s on a single GPU.

\paragraph{Attention type}
The classical softmax-based attention mechanism (such as that used in GAT) is a \textit{sparse} attention where most of the importance is on the maximal value.
In contrast, the sigmoid-based edge gating mechanism for our graph ConvNet (Eq. \eqref{eqn:gcn-node}) can be termed as a \textit{dense} attention mechanism, where all saturated sigmoids lead to equal importance.
We briefly experimented with both sparse and dense attention mechanisms for TSP and found dense attention to lead to marginally better performance and lesser GPU memory consumption.
Figure \ref{fig:val-opt-gap-attn} displays the validation optimality gap vs. number of training samples of TSP50 for both attention types (all other model hyperparameters are the same).

\section{Comments on Supervised Learning vs Reinforcement Learning}
\label{sl-vs-rl}

As noted in \cite{bengio2018machine}, the performance of supervised learning-based models for combinatorial optimization problems depends on the availability of a large set of optimal or high-quality solutions. 
Thus, two key issues arise when formulating these problems as supervised learning tasks: 
(1) we are restricted to learning well-studied problems for which optimal solvers or high-quality heuristic algorithms are available; and
(2) we can only train on small-scale problem sizes as it is intractable to build datasets for large instances of NP-hard problems. 

Although reinforcement learning is known to be more computationally expensive/less sample efficient than supervised learning, it does not require the generation of pairs of problem instances and solutions. 
As long as a problem can be formulated via a reward signal for making sequential decisions, a policy can be trained via RL.
Hence, most recent work on learning-based approaches for TSP have used RL \citep{deudon2018learning,kool2018attention}.
Comparatively poor performance of SL methods \citep{vinyals2015pointer,nowak2017note} have supported the argument in favour of reinforcement learning.

Unlike \cite{vinyals2015pointer} and \cite{nowak2017note}, 
our approach uses deep graph ConvNets which are able to learn from a larger training set of optimal TSP solutions (one million instances). 
Our approach outperforms all other learning-based approaches in terms of both solution quality and sample efficiency. 
This result does not come as a surprise as SL techniques usually outperform RL techniques given sufficient amount of training data. 
However, the advantage of SL quickly diminishes for larger instances. Generating one million training samples for problem sizes beyond hundreds of nodes can become intractable in terms of computation and speed. 
The rapid increase in combinatorial complexity of TSP as problem size increases, termed as \textit{combinatorial explosion}, makes it intractable to scale our approach to large TSPs.

Thus, incorporating RL to tackle arbitrary problem sizes is the next natural development for our approach:
Future work shall explore learning a policy network by graph ConvNet, optimizing the tour length and applying beam search without optimal solutions.
Supervised training on small instances and transfer learning by fine-tuning model parameters on large instances using RL is an attractive approach for scaling up to realistic sizes beyond hundreds of nodes.

\section{Solution Visualizations}
\label{viz}

\begin{figure}[h!]
    \subfloat[This instance can be considered easy because the nodes are evenly distributed and the optimal tour is simply the cycle enclosing all the nodes.
    The model is able to confidently predict the groundtruth edges, resulting in greedy search being able to find the optimal tour easily.]{
	    \begin{minipage}[c]{\textwidth}
	        \centering
	        \includegraphics[width=\textwidth]{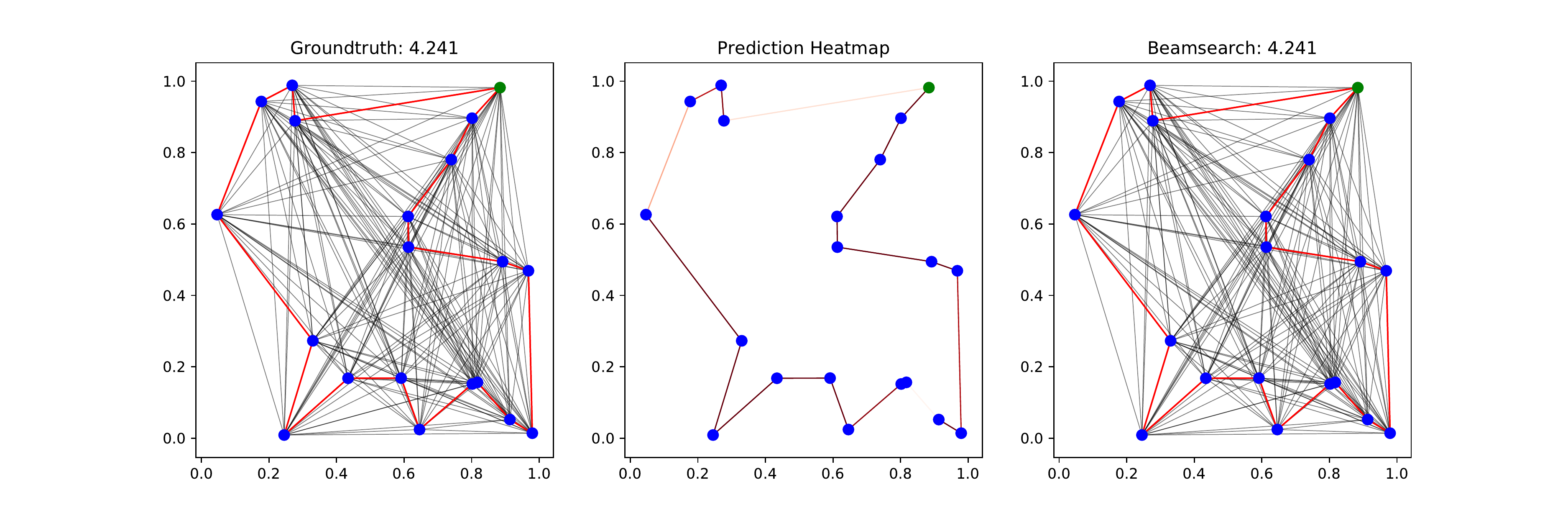}
	    \end{minipage}}
    \newline
    \subfloat[This instance is comparatively harder due to the presence of a cluster of nodes in the bottom-center.
    The model predicts several possible edges at the bottom-center and the exact contours are then found using beam search.]{
	    \begin{minipage}[c]{\textwidth}
	        \centering
	        \includegraphics[width=\textwidth]{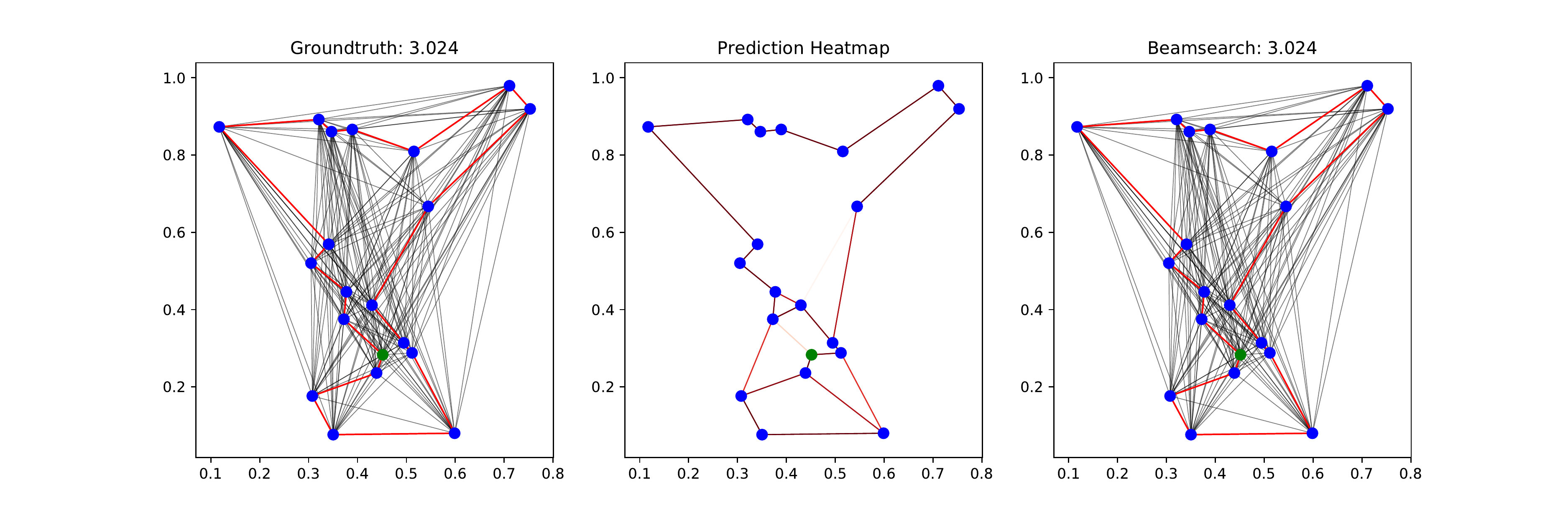}
	    \end{minipage}}
    \newline
    \subfloat[The model predicts a shorter tour than Concorde for this instance by choosing a different contour around the nodes at the top-left of the graph.]{
	    \begin{minipage}[c]{\textwidth}
	        \centering
	        \includegraphics[width=\textwidth]{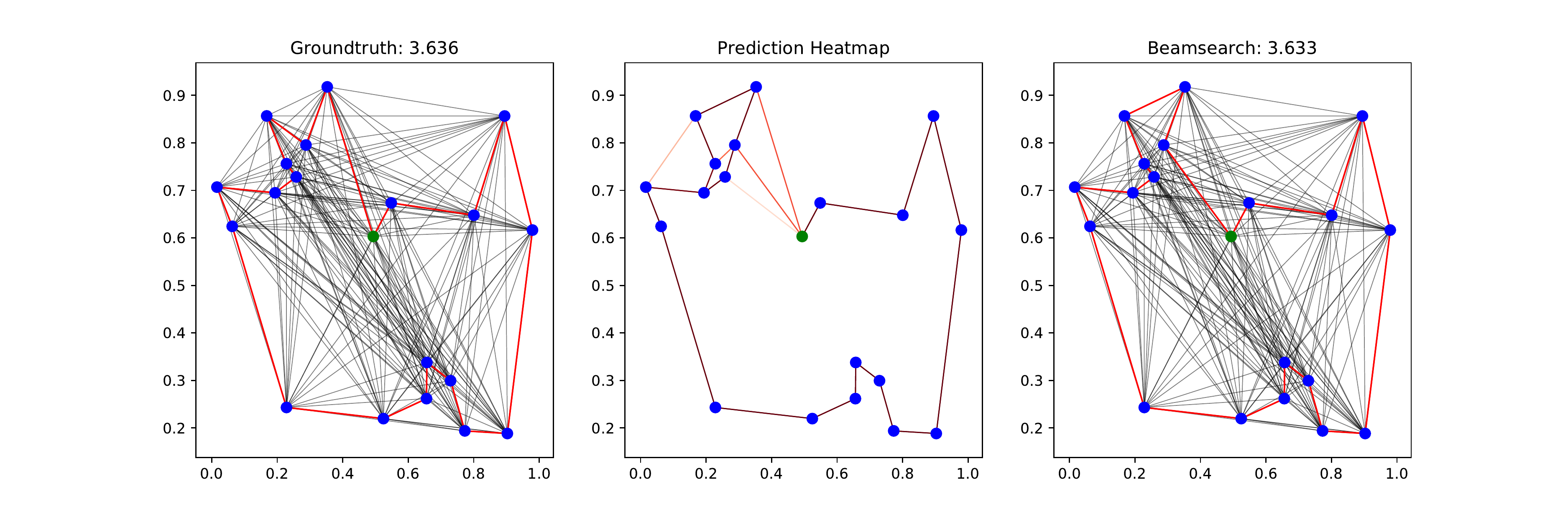}
	    \end{minipage}}
\caption{TSP20 model prediction visualization for sample test set instances.}
\label{fig:tsp20}
\end{figure}

\begin{figure}[h!]
    \subfloat[As the complexity of problem instances increases, multiple connections are predicted for nodes lying in dense clusters. The model relies on beam search to find the optimal tour.]{
	    \begin{minipage}[c]{\textwidth}
	        \centering
	        \includegraphics[width=\textwidth]{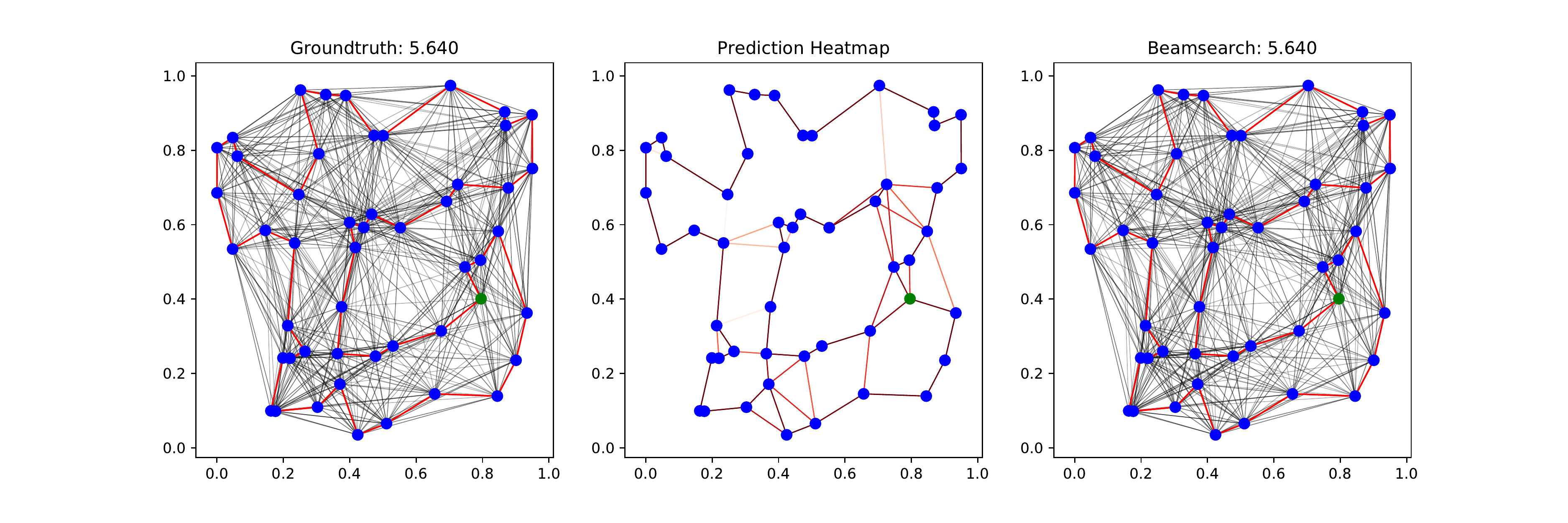}
	    \end{minipage}}
    \newline
    \subfloat[For this instance, the model predicts a tour that makes different envelopes around the nodes at the top-left and top-right corners compared to the groundtruth. However, the final solution is very close to optimal in terms of length.]{
	    \begin{minipage}[c]{\textwidth}
	        \centering
	        \includegraphics[width=\textwidth]{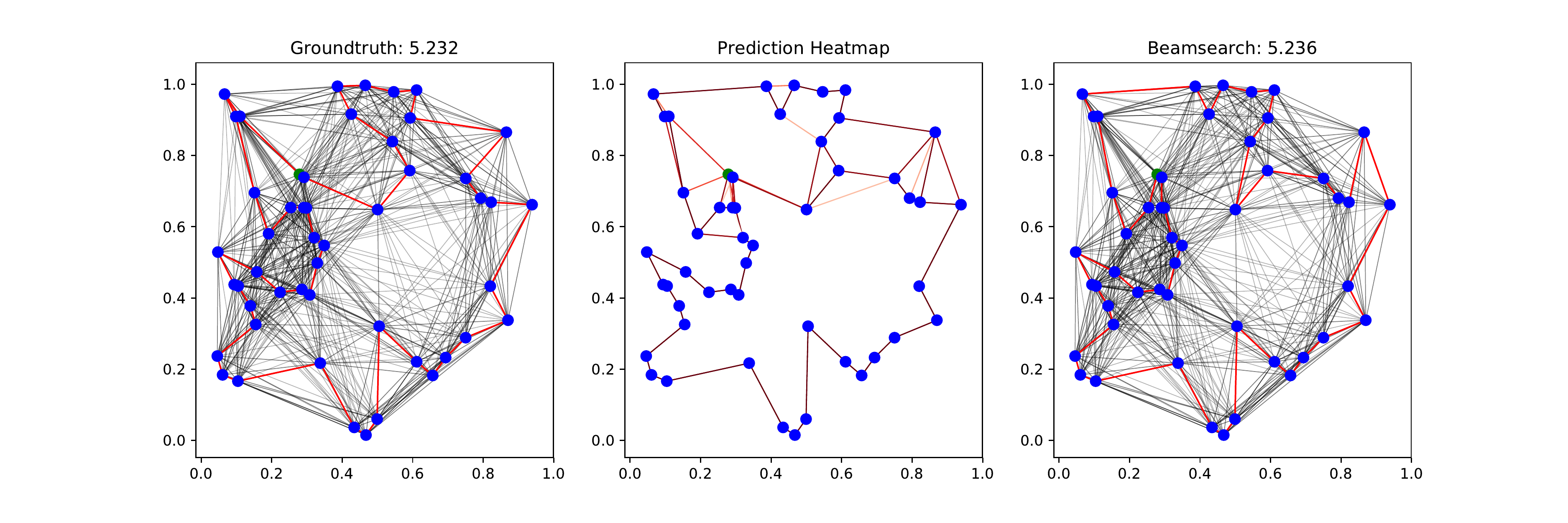}
	    \end{minipage}}
    \newline
    \subfloat[Most of the complexity of this instance comes from the cluster of nodes at the bottom-right corner. The model predicts a shorter tour than Concorde by choosing a different contour around the nodes at this corner.]{
	    \begin{minipage}[c]{\textwidth}
	        \centering
	        \includegraphics[width=\textwidth]{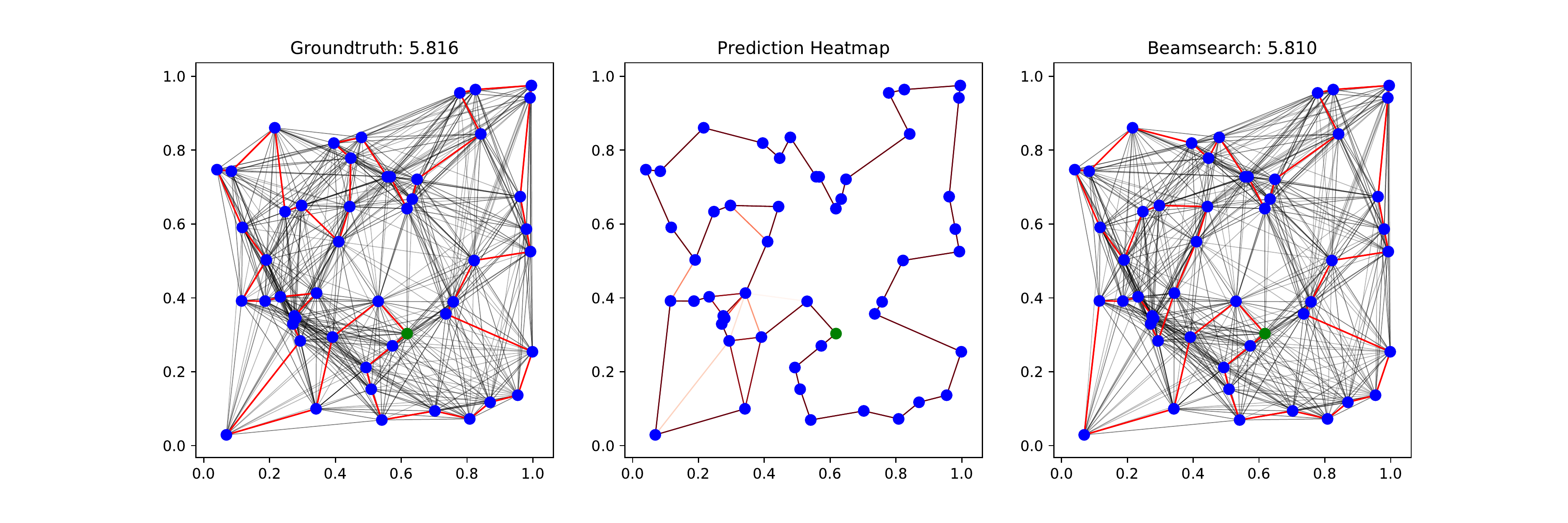}
	    \end{minipage}}
\caption{TSP50 model prediction visualization for sample test set instances.}
\label{fig:tsp50}
\end{figure}

\begin{figure}[h!]
    \subfloat[The complexity of this instance is due to the envelopes from the top-center into the center of the graph in the optimal tour. The model relies on beam search to navigate the complexity and predicts the optimal tour.]{
	    \begin{minipage}[c]{\textwidth}
	        \centering
	        \includegraphics[width=\textwidth]{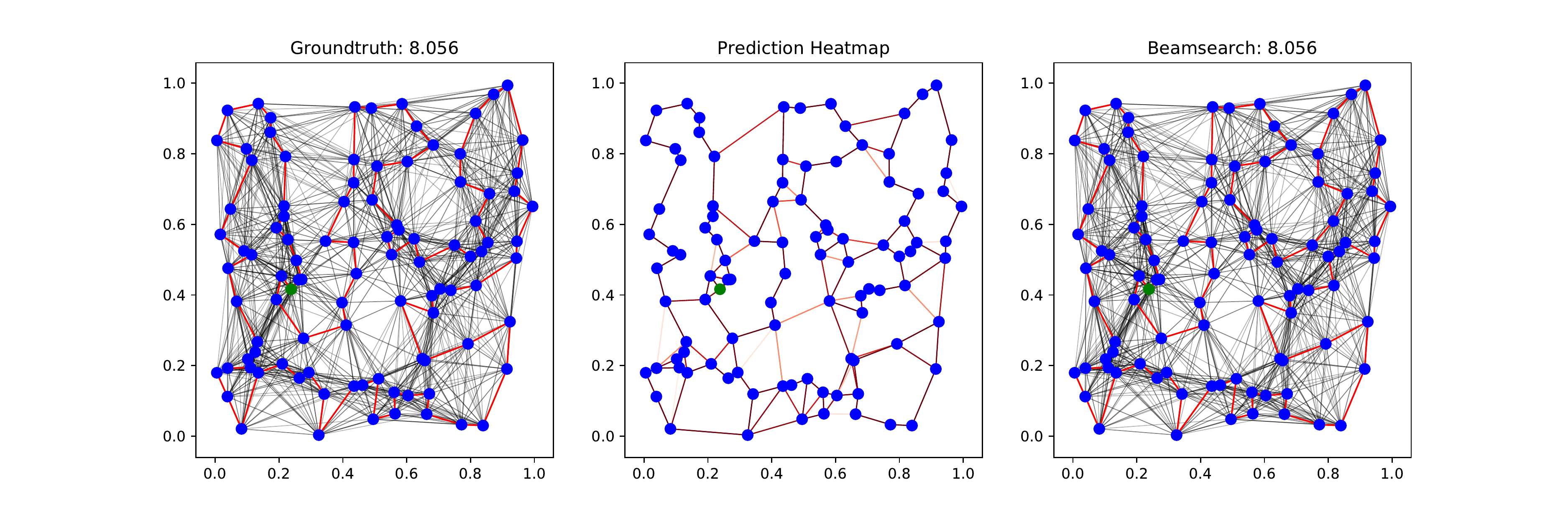}
	    \end{minipage}}
    \newline
    \subfloat[This instance is harder than (a) because there several envelopes towards the center of the graph in the optimal tour. The model is unable to confidently identify these envelopes and beam search is unable to find the optimal contours around the center. This is reflected in extremely long connections being predicted for certain nodes.]{
	    \begin{minipage}[c]{\textwidth}
	        \centering
	        \includegraphics[width=\textwidth]{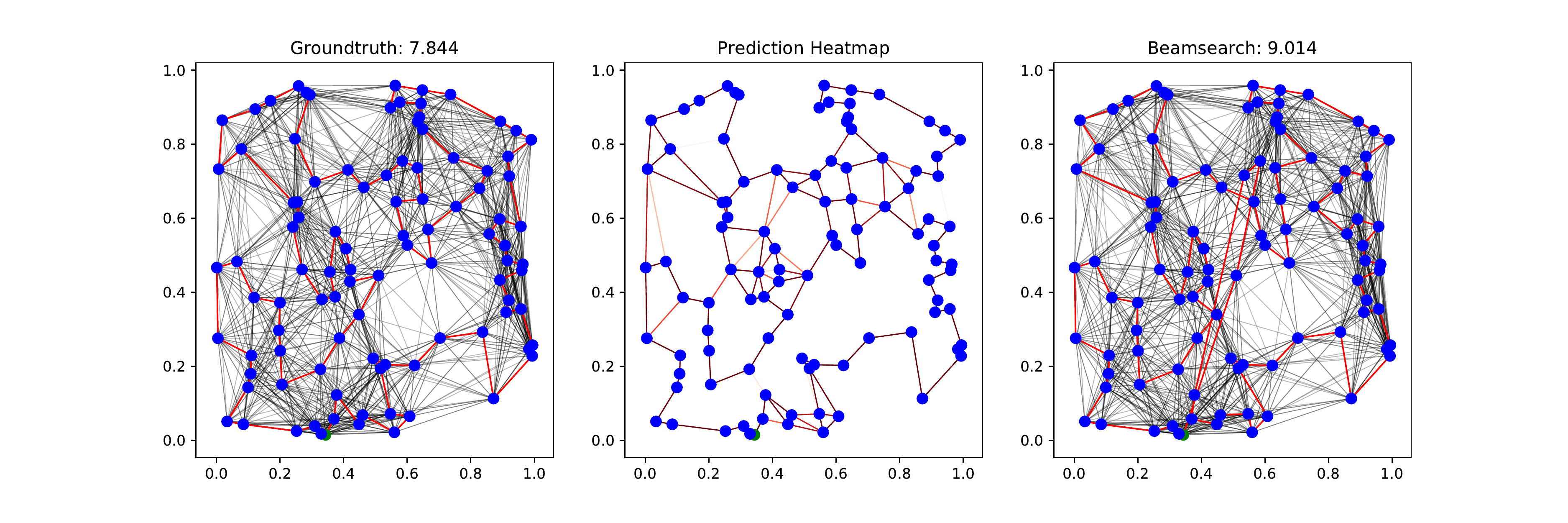}
	    \end{minipage}}
    \newline
    \subfloat[The model solves this instance better than Concorde by choosing different contours around the nodes at the top-right corner of the graph.]{
	    \begin{minipage}[c]{\textwidth}
	        \centering
	        \includegraphics[width=\textwidth]{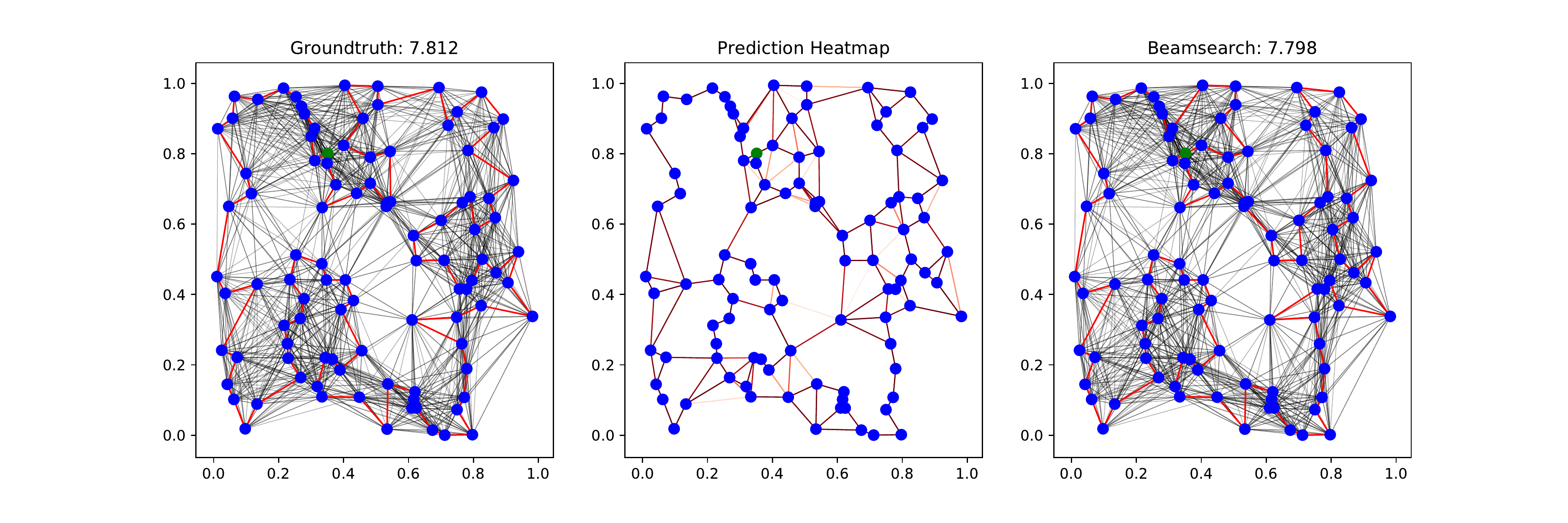}
	    \end{minipage}}
\caption{TSP100 model prediction visualization for sample test set instances.}
\label{fig:tsp100}
\end{figure}

Figures \ref{fig:tsp20}, \ref{fig:tsp50} and \ref{fig:tsp100}
display prediction visualizations for samples from test sets of various problem instances.
In each figure, the first panel shows the input $k$-nearest neighbor graph and the groundtruth TSP tour.
The second panel represents the probabilistic heat-map output of the graph ConvNet model.
The final panel shows the predicted TSP tour after a beam search procedure on the heat-map.

For small instances where the nodes are evenly distributed, the model is able to confidently identify most of the tour edges in the heat-map, resulting in greedy search being able to find close to optimal tours.
As instance size increases, the prediction heat-map reflects the \emph{combinatorial explosion} in TSP
and beam search is essential for finding the optimal tour.

\end{document}